\documentclass[pdflatex,sn-mathphys-num]{sn-jnl}

\usepackage{xr}
\makeatletter
\newcommand*{\addFileDependency}[1]{
  \typeout{(#1)}
  \@addtofilelist{#1}
  \IfFileExists{#1}{}{\typeout{No file #1.}}
}
\makeatother

\newcommand*{\myexternaldocument}[1]{
    \externaldocument{#1}
    \addFileDependency{#1.tex}
    \addFileDependency{#1.aux}
}

\myexternaldocument{SI}

\usepackage{graphicx}%
\usepackage{multirow}%
\usepackage{amsmath,amssymb,amsfonts}%
\usepackage{amsthm}%
\usepackage{mathrsfs}%
\usepackage[title]{appendix}%
\usepackage{xcolor}%
\usepackage{textcomp}%
\usepackage{manyfoot}%
\usepackage{booktabs}
\usepackage{algorithm}%
\usepackage{algorithmicx}%
\usepackage{algpseudocode}%
\usepackage{listings}%
\usepackage{float}%
\usepackage{array}%
\usepackage{comment}%
\usepackage{subcaption}%
\usepackage{hyperref}%
\usepackage[version=4]{mhchem} 
\usepackage{tabularx}
\usepackage{multirow}
\usepackage{subcaption}

\theoremstyle{thmstyleone}%
\theoremstyle{thmstyletwo}%

\theoremstyle{thmstylethree}%

\begin{document}
\title[Article Title]{Interpolation-Driven Machine Learning Approaches for Plume Shine Dose Estimation: A Comparison of XGBoost, Random Forest, and TabNet}


\author*[1,2]{\fnm{Biswajit} \sur{Sadhu}}\email{bsadhu@barc.gov.in, biswajit.chem001@gmail.com}

\author[2,4]{\fnm{Kalpak} \sur{Gupte}}

\author[3]{\fnm{Trijit} \sur{Sadhu}}

\author[1,2]{\fnm{S} \sur{Anand}}

\affil[1]{\orgdiv{Health Safety \& Environment Group}, \orgname{Bhabha Atomic Research Centre},  \state{Mumbai}, \postcode{400085}, \country{India}}

\affil[2]{\orgdiv{Homi Bhabha National Institute}, \state{Mumbai}, \postcode{400094},  \country{India}}

\affil[3]{\orgdiv{Birla Institute of Technology And Science}, \orgaddress{\street{Street}, \city{PILANI}, \postcode{333031}, \state{Rajasthan}, \country{India}}}

\affil[4]{\orgdiv{Department of Computer Engineering \& Technology}, \orgname{Dr. Vishwanath Karad MIT World Peace University}, \state{Pune}, \postcode{411038}, \country{India}}

\abstract{Despite the success of machine learning (ML) in surrogate modeling, its use in radiation dose assessment is limited by safety-critical constraints, scarce training-ready data, and challenges in selecting suitable architectures for physics-dominated systems. Within this context, rapid and accurate plume shine dose estimation serves as a practical test case, as it is critical for nuclear facility safety assessment and radiological emergency response, while conventional photon-transport-based calculations remain computationally expensive. In this work, an interpolation-assisted ML framework was developed using discrete dose datasets generated with the pyDOSEIA suite for 17 gamma-emitting radionuclides across varying downwind distances, release heights, and atmospheric stability categories. The datasets were augmented using shape-preserving interpolation to construct dense, high-resolution training data. Two tree-based ML models (Random Forest and XGBoost) and one deep learning (DL) model (TabNet) were evaluated to examine predictive performance and sensitivity to dataset resolution. All models showed higher prediction accuracy with the interpolated high-resolution dataset than with the discrete data; however, XGBoost consistently achieved the highest accuracy. Interpretability analysis using permutation importance (tree-based models) and attention-based feature attribution (TabNet) revealed that performance differences stem from how the models utilize input features. Tree-based models focus mainly on dominant geometry–dispersion features (release height, stability category, and downwind distance), treating radionuclide identity as a secondary input, whereas TabNet distributes attention more broadly across multiple variables. For practical deployment, a web-based GUI was developed for interactive scenario evaluation and transparent comparison with photon-transport reference calculations.}


\maketitle

\section{Introduction}\label{sec1}

Machine learning (ML) is increasingly used to accelerate scientific modeling by enabling fast and accurate surrogate predictions for complex physical processes in the domain of climate science, materials engineering, and medical imaging. However, the adoption of ML in radiation exposure and dose assessment still remains rather limited.\cite{sadhu2025raddqn, luo2020machine} This is mainly due to the safety‐critical nature of radiological applications, where even small prediction errors can have serious consequences, and to the lack of training-ready dense datasets.

\begin{figure}[htbp]
\centering
\includegraphics[width=0.9\linewidth]{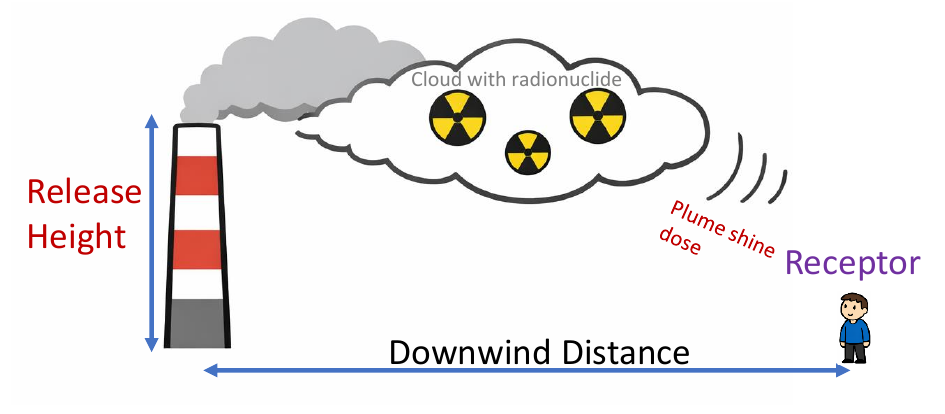}
\caption{Schematic representation of plume shine exposure geometry, illustrating radionuclide release from an elevated source, atmospheric transport of the radioactive cloud, and external gamma dose received by a downwind receptor as a function of release height and downwind distance. Although not explicitly shown, the atmospheric stability category governs plume dispersion characteristics and significantly influences the resulting plume shine dose.}
\label{fig:psd_image}
\end{figure}

Plume shine (also known as cloud shine) is a representative example of this challenge. It refers to the external radiation dose received from gamma emissions of an airborne radioactive cloud following a nuclear or radiological release (Fig: \ref{fig:psd_image}). Reliable and timely estimation of plume shine dose is of paramount interest both for emergency preparedness and response, supporting decisions on evacuation and sheltering \cite{mckenna2000protective}, and for the safety assessment of new facilities, where it plays a key role in dose apportionment and regulatory compliance. Modeling based approaches are therefore a stndard component of plume shine estimation by the regulatory authorities and emergency planners \cite{armand2005simulation, satoh2021simulation, karmakar2022development, accuracycloudshine2017}.

Traditionally, plume shine dose is computed by coupling atmospheric dispersion models with photon transport and dose conversion formalisms. This approach typically involves a triple integration over the spatial extent of the plume, the gamma‐ray energy spectrum of the radionuclide inventory, attenuation and buildup factors that account for air absorption and scattering. \textit{While physically rigorous, these approaches incur substantial computational cost, limiting their applicability in time‐critical operational settings, especially when assessments involve multiple radionuclides with complex gamma spectra or require fine spatial resolution} \cite{accuracycloudshine2017}. To reduce computational burden, many assessment frameworks use simplified transport representations such as Gaussian plume models \cite{Cember2009HealthPhysics, Till2008RadiologicalRisk}. In this context, our previous work introduced pyDOSEIA, an open‐source Python package that implements Gaussian plume dispersion for comprehensive radiological impact assessment, including plume shine dose along with other exposure pathways \cite{Sadhu2026pyDOSEIA}. Nevertheless, analytical plume models remain constrained by assumptions of steady meteorology and simple terrain, while high‐fidelity physics‐based simulations remain too slow for real‐time decision support \cite{armand2021topical}.

These limitations motivate the development of ML‐based surrogate models for plume shine dose prediction. Once trained, such models can provide near‐instantaneous estimates, enabling rapid scenario screening and real‐time support during radiological emergencies. However, applying ML in this domain presents challenges that differ from those in many standard ML applications. Plume shine dose data are usually available only as sparse tables defined for a limited number of combinations of downwind distance, release height, and atmospheric stability class. Training ML models directly on such discontinuous data often results in unstable predictions and poor generalization, which is unacceptable in safety‐critical applications where reliability and physical consistency are essential.

In addition to data limitations, the choice of ML model plays a critical role. For structured tabular data, tree‐based ensemble methods such as Random Forest (RF) \cite{breiman2001random} and XGBoost \cite{chen2016xgboost} have become the standard in practical ML workflows. They are widely used in industry and research and have consistently achieved strong performance in data‐science competitions. Their success is largely due to their robustness, ability to model nonlinear relationships, and built‐in feature selection that reduces the influence of weak or irrelevant variables. On the contrary, deep learning (DL) based approaches have traditionally been reported to be less effective on tabular datasets, although recent architectures such as TabNet \cite{arik2021tabnet} have been proposed to address this gap through incorporation of sequential attention mechanisms. Despite this, several systematic studies have shown the superiority of ensemble methods over DL models for tabular prediction tasks, particularly when datasets are limited and strongly structured \cite{fayaz2022deeptabular, shwartz2022tabular}. Moreover, understanding the underlying reasons for these performance differences remains an active area of research. This leads to an important practical question for radiation modeling:
\emph{Should plume shine dose prediction rely on well-established ensemble methods, or can newer attention-based deep architectures provide clear advantages in this safety-critical domain?}

In this work, we address both data and model selection challenges and propose an interpolation‐assisted ML framework for plume shine dose prediction. \textit{The study is guided by two key objectives: achieving high predictive accuracy and ensuring transparent, model‐appropriate interpretability.} To this end, we conduct a systematic comparison of three leading approaches for tabular data -- RF, XGBoost, and TabNet -- within a unified and controlled experimental setting.

Our framework is based on three key methodological elements.
First, in order to address the limitations of sparse and discretized dose tables, we employ data augmentation using Piecewise Cubic Hermite Interpolating Polynomials (PCHIP) \cite{fritsch1980monotone} that preserves the underlying physical trends and therefore enabling the construction of high-quality high-resolution training datasets.  
Second, we conduct a systematic comparison of model inductive biases by benchmarking established tree-based ensemble methods against DL-based TabNet.
Third, we adopt an architecture-aware interpretability strategy by applying conditional permutation importance to the ensemble models and analyzing feature relevance in TabNet through its intrinsic attention mechanisms. This approach enables a consistent examination of how different models utilize input features and therefore helps understanding the root cause of model's performance behavior, in line with established practices in interpretable machine learning \cite{molnar2019interpretable}.

Finally, to improve transparency and practical usability, we include an interactive Streamlit-based interface that allows real-time comparison between machine-learning predictions and physics-based reference calculations, helping connect the methodological analysis with practical, scenario-based insight.

\section{Analytical Plume Shine Dose Computation and Methodological Motivation}

\label{sec:plumeshine_method}

Plume shine dose refers to the external radiation exposure arising from photon emissions of an airborne radioactive cloud incident at a receptor location. In radiation protection practice, plume shine dose is commonly evaluated using analytical point‐kernel formulations that integrate photon transport contributions over the full spatial extent of the radioactive plume. These formulations explicitly account for source geometry, atmospheric attenuation, and photon scattering, and are widely regarded as the reference approach for high‐fidelity dose estimation.

For a receptor located at $(x_1, y_1, z_1)$, the plume shine dose rate can be expressed using a point‐kernel formulation as
\begin{equation}
\dot{D} = \iiint \alpha E_\gamma
\frac{\mu_a}{\rho}
\frac{B(\mu r)}{4\pi r^2}
\exp(-\mu r)\,
\chi(x,y,z)\,
\mathrm{d}x\,\mathrm{d}y\,\mathrm{d}z
\label{ps_eq}
\end{equation}

where $r=\sqrt{(x-x_1)^2+(y-y_1)^2+(z-z_1)^2}$ is the source–receptor distance, $E_\gamma$ denotes the photon energy, $\mu$ and $\mu_a$ are the linear attenuation and energy‐absorption coefficients in air, $\rho$ is the air density, $B(\mu r)$ is the buildup factor accounting for scattered photons, $\chi(x,y,z)$ is the concentration of radionuclide in plume, and $\alpha$ is a unit conversion constant.

As evident from Eq.~(\ref{ps_eq}), plume shine dose estimation involves a three‐dimensional spatial integration, combined with a summation over the discrete gamma energies emitted by each radionuclide. The effective spatial extent of this integration is not uniform but is governed by atmospheric dispersion and photon attenuation characteristics. In the lateral (y) and vertical (z) directions, the integration limits are determined by the plume spread parameters $\sigma_y$ and $\sigma_z$, which depend strongly on the atmospheric stability category\cite{pasquill1961estimation}. Unstable conditions (stability classes A--C) are associated with enhanced turbulent mixing, leading to broader plume dimensions, whereas stable conditions (classes E--F) suppress dispersion and confine the plume to a narrower region with limited vertical extent. Neutral conditions (class D) represent an intermediate regime between these extremes. Along the downwind direction (x), the effective integration range is constrained by the photon mean free path (MFP), beyond which dose contributions diminish rapidly due to exponential attenuation. Together, these stability- and photon-based constraints define the spatial domain contributing meaningfully to plume-shine dose. \textit{In practical implementations, the integral does not have a closed‐form solution and is therefore evaluated numerically using quadrature or discretization‐based techniques}.\cite{hukkoo1988manual}

\subsection{Numerical and Computational Challenges}

Despite its strong physical basis as described above, analytical plume shine dose computation involves several numerical and computational difficulties that restrict its routine use in operational and large-scale applications:

\begin{itemize}
\item \textbf{High-dimensional integration:}
The spatial integration extends over large downwind, crosswind, and vertical domains. Even when the integration region is truncated to a few photon mean free paths \cite{pecha2014unconventional}, the associated computational cost remains significant.

\item \textbf{Near-field numerical stiffness:}
The presence of the $1/r^2$ kernel introduces steep gradients in the vicinity of the receptor, which necessitates fine spatial discretization to maintain numerical stability and acceptable accuracy.

\item \textbf{Multiple gamma emissions:}
Many radionuclides emit multiple gamma lines (e.g. \ce{^{154}Eu}), each requiring separate attenuation and buildup calculations. As shown in our earlier implementation using \texttt{pyDOSEIA} \cite{Sadhu2026pyDOSEIA}, the computational cost increases approximately linearly with the number of gamma energies considered.

\item \textbf{Scaling with resolution and scenario count:}
Increasing spatial resolution, downwind distance, or release height expands the effective integration volume. In addition, realistic consequence assessment requires repeated evaluations across numerous meteorological and source scenarios. Together, these factors lead to a rapid growth in computational demand.
\end{itemize}

Taken together, these limitations make detailed plume shine dose calculations computationally intensive and difficult to apply in contexts that require rapid turnaround, such as emergency response and large-scale scenario screening.

\subsection{Motivation for Data‐Driven Surrogate Modeling}

The computational challenges outlined above motivate the use of data-driven surrogate models for plume shine dose estimation. Instead of replacing the underlying physics-based formulation, machine learning models can be trained on reference datasets generated through analytical–numerical calculations and then employed to deliver rapid predictions across the relevant input parameter space.

However, developing reliable surrogate models for plume shine dose is not straightforward. A major bottleneck lies in the dataset itself, which, even when generated using analytical calculations, is typically sparse and discretized due to the combinatorial combinations of downwind distance, release height, and atmospheric stability category. As shown later in this study, training machine learning models directly on such discontinuous data can result in unstable predictions and limited generalization capability, which is particularly problematic in safety-critical applications.

To address this limitation, the present study adopts an interpolation‐assisted dataset construction strategy, in which physically consistent interpolation methods are used to transform sparse analytical dose tables into smooth, high‐resolution datasets suitable for modern ML. This strategy forms the methodological bridge between classical radiation transport modeling and the ML framework introduced in the following sections.

\section{Data Generation and Preparation for ML Modeling}
\label{sec:data_gen}

\subsection{Low-Resolution Dataset Generation using pyDOSEIA}
\label{sec:dlr}

The base dataset used in this study was generated using the \texttt{pyDOSEIA} package \cite{Sadhu2026pyDOSEIA}, which implements a Gaussian plume dispersion model for radiological impact assessment, including plume shine dose estimation. \textit{The simulations assume a short-term plume release with a unit source term ($Q = 1$ Bq s$^{-1}$) and a reference wind speed of $U = 1$ m s$^{-1}$ for each radionuclide, evaluated across different release heights and downwind distances}. This initial dataset, hereafter referred to as the low-resolution dataset (\textbf{$\mathcal{D}_{\mathrm{LR}}$}), consists of plume shine dose values computed at discrete combinations of physical and meteorological parameters.

Specifically, the dataset includes 17 gamma-emitting radionuclides: \ce{^{137}Cs}, \ce{^{134}Cs}, \ce{^{41}Ar}, \ce{^{135}Xe}, \ce{^{60}Co}, \ce{^{131}I}, \ce{^{132}I}, \ce{^{87}Kr}, \ce{^{88}Kr}, \ce{^{85}Kr}, \ce{^{85}Sr}, \ce{^{103}Ru}, \ce{^{106}Ru}, \ce{^{22}Na}, \ce{^{152}Eu}, \ce{^{154}Eu}, and \ce{^{155}Eu}.

These radionuclides were selected to represent a broad and realistic range of plume shine contributors encountered in nuclear and radiological applications. The set includes key fission products released during reactor accidents and fuel handling events (e.g., Cs, I, and Ru isotopes), activation products relevant to routine operations and research facilities (e.g., \ce{^{41}Ar}, \ce{^{60}Co}, \ce{^{22}Na}), and noble gases that dominate early plume shine immediately after release because of their high mobility and limited deposition (e.g., Xe and Kr isotopes). The inclusion of multiple europium isotopes allows the dataset to represent radionuclides with more complex gamma emission spectra and longer half-lives, which are particularly relevant for facility characterization and long-term dose budgeting. Collectively, the selected radionuclides span a broad range of half-lives, gamma energies, emission multiplicities, and chemical properties, providing a representative testbed for evaluating plume shine dose modeling and machine-learning-based surrogate approaches.

The resulting dataset consists of four key input features and one target variable, as summarized in Table~\ref{tab:dataset_desc}. The input features include radionuclide identity, atmospheric stability category, release height, and downwind distance, while the target variable is the corresponding plume shine dose at the receptor location (Fig. \ref{fig:psd_image}).

\begin{table}[h]
\centering
\caption{Description of dataset features used for plume shine dose prediction}
\label{tab:dataset_desc}
\begin{tabular}{p{3cm} c p{3.2cm} p{5.2cm}}
\hline
\textbf{Feature} & \textbf{Type} & \textbf{Range / Categories} & \textbf{Description} \\ \hline
Radionuclide & Categorical & 17 gamma emitters &
Gamma-emitting radionuclide (e.g., \ce{^{137}Cs}, \ce{^{41}Ar}, \ce{^{131}I}, \ce{^{135}Xe}). \\ \hline
Atmospheric Stability Category & Categorical & A--F & Pasquill--Gifford atmospheric stability class. \\ \hline

Release Height & Numerical & 10--200 m (step: 10 m) &
Release height above ground level. \\ \hline
Downwind Distance & Numerical & 25--2000 m &
Downwind distance along plume centerline. \\ \hline
Plume Shine Dose & Continuous (Target) & $\sim10^{-13}$--$10^{-8}$ &
External gamma dose at receptor ($\mu$Sv/hr). \\ \hline
\end{tabular}
\end{table}

\begin{figure}[H]
\centering
\includegraphics[width=0.9\linewidth]{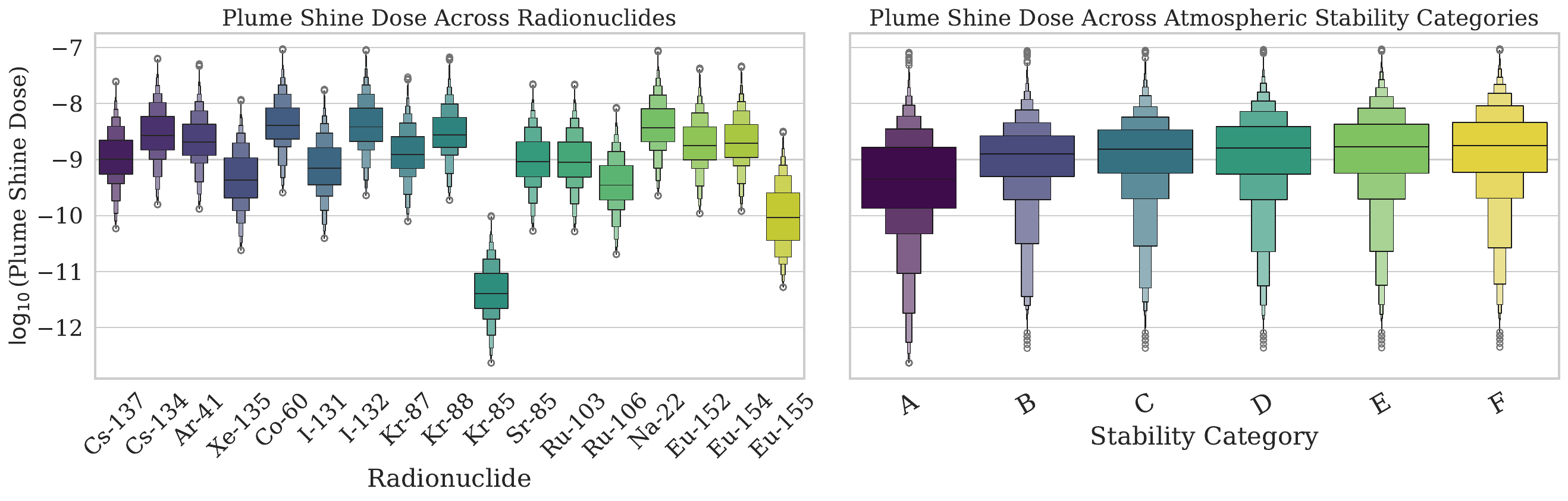}
\caption{Distribution of log$_{10}$ plume shine dose across radionuclides (left) and atmospheric stability categories A–F (right). The boxen plots show median, spread, and tail behavior, with stability classes A–F representing Pasquill–Gifford regimes from unstable to stable}
\label{fig:Catwise_violin}
\end{figure}

The plume shine dose spans several orders of magnitude, reflecting variations in radionuclide gamma yield, atmospheric dispersion conditions, and source–receptor geometry. Figure~\ref{fig:Catwise_violin} presents the distribution of dose values across categorical features on a logarithmic scale, illustrating both central tendencies and variability across radionuclides and stability classes.

Although $\mathcal{D}_{\mathrm{LR}}$ is physically consistent, its discrete structure restricts its effectiveness for training data-driven surrogate models, which typically require quasi-continuous coverage of the input domain. The impact of this discretization on predictive performance is analyzed later in this study. This limitation motivates the interpolation strategy presented in the following subsection.

\subsection{High-Resolution Dataset Construction via Stepwise Interpolation}
\label{sec:interp_method}

To convert the sparse low-resolution into the high-resolution dataset, an interpolation-based densification was applied only along the downwind distance, which is irregularly and sparsely discretized in the original data. Radionuclide identity and atmospheric stability category were preserved as categorical variables. Unlike offline synthetic data augmentation frameworks (e.g., SynthCity) that generate distribution-expanding samples\cite{qian2023synthcity}, the present approach performs deterministic densification without introducing synthetic variability beyond the original discrete data manifold.

\paragraph{Distance-wise interpolation using PCHIP.}
One-dimensional interpolation was performed along the downwind distance axis for each unique combination of radionuclide, release height, and atmospheric stability category. The PCHIP method was chosen because it preserves shape and monotonic trends, which are key characteristics of plume shine dose variation with downwind distance and release height.

\begin{figure}
\centering
\includegraphics[width=0.9\textwidth]{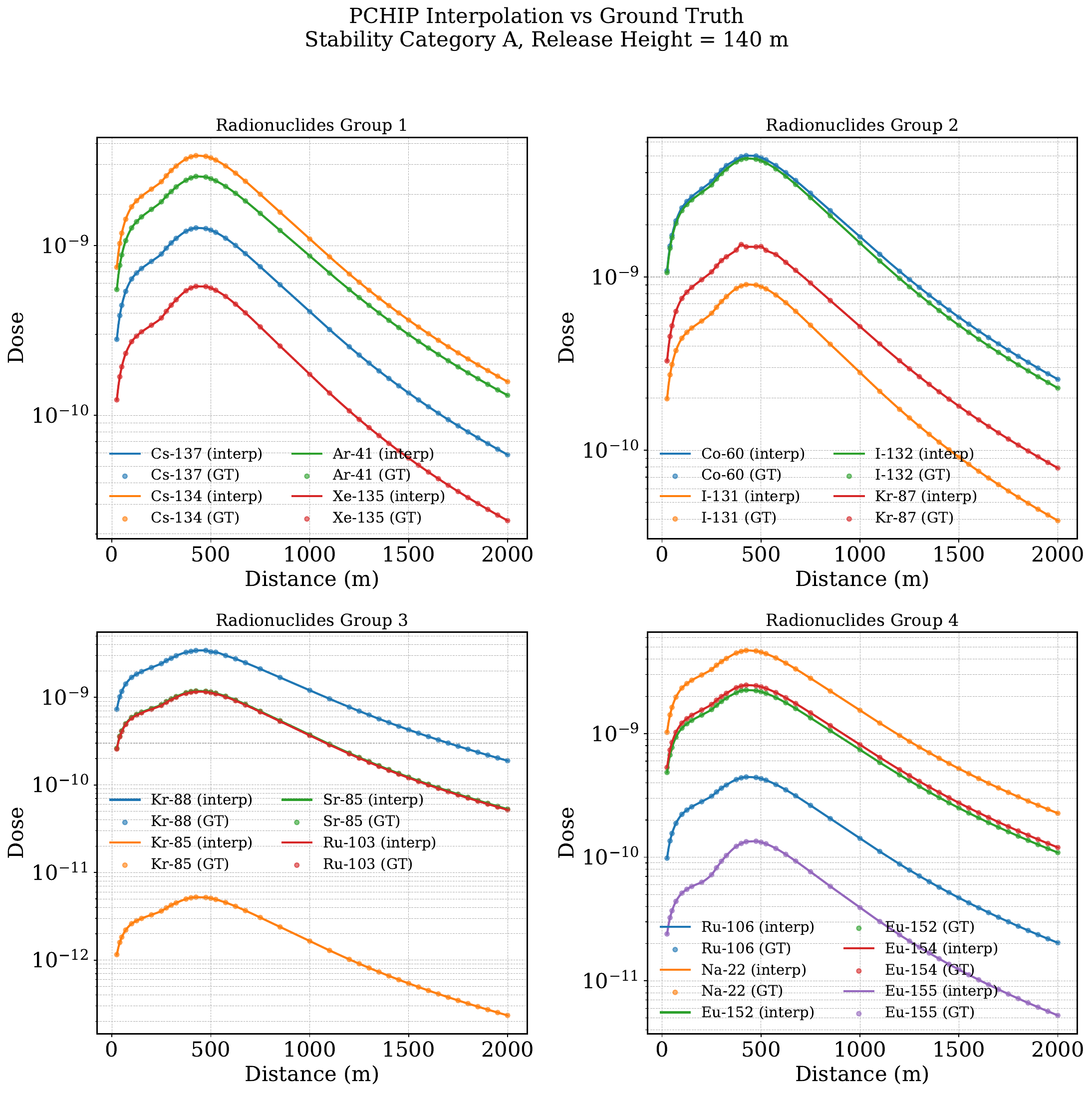}
\caption{Distance-wise comparison between PCHIP-interpolated plume shine dose and numerically calculated ground-truth values for representative radionuclides under stability category A at a release height of 140 m. The agreement demonstrates the shape-preserving and monotonic behavior of the interpolation method.}
\label{fig:1dphcip_a}
\end{figure}

\begin{figure}
\centering
\includegraphics[width=0.9\textwidth]{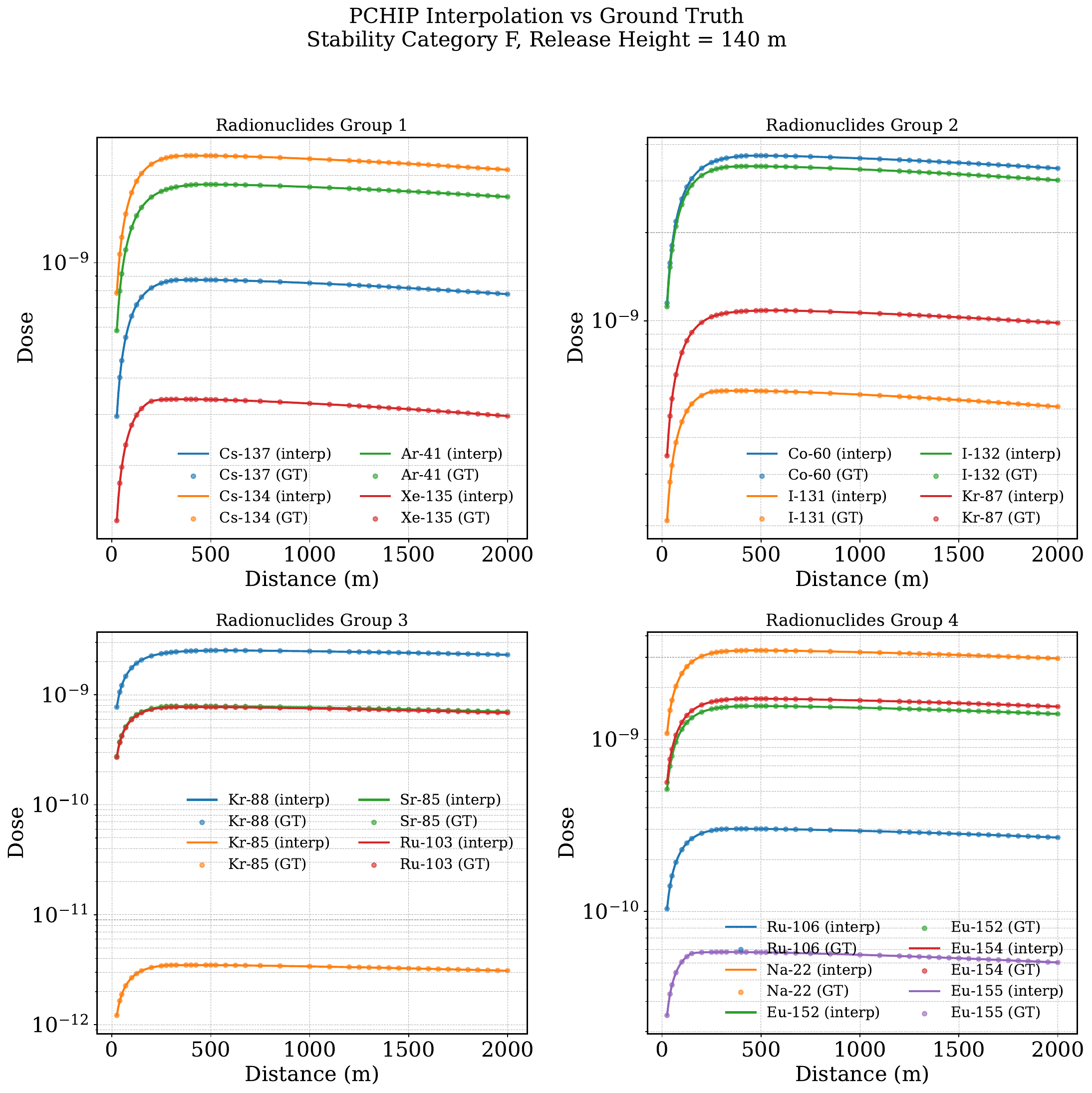}
\caption{Distance-wise comparison between PCHIP-interpolated plume shine dose and numerically calculated ground-truth values for representative radionuclides under stability category F at a release height of 140 m. The agreement demonstrates the shape-preserving and monotonic behavior of the interpolation method.}
\label{fig:1dphcip_f}
\end{figure}

\begin{figure}
\centering
\includegraphics[width=0.9\textwidth]{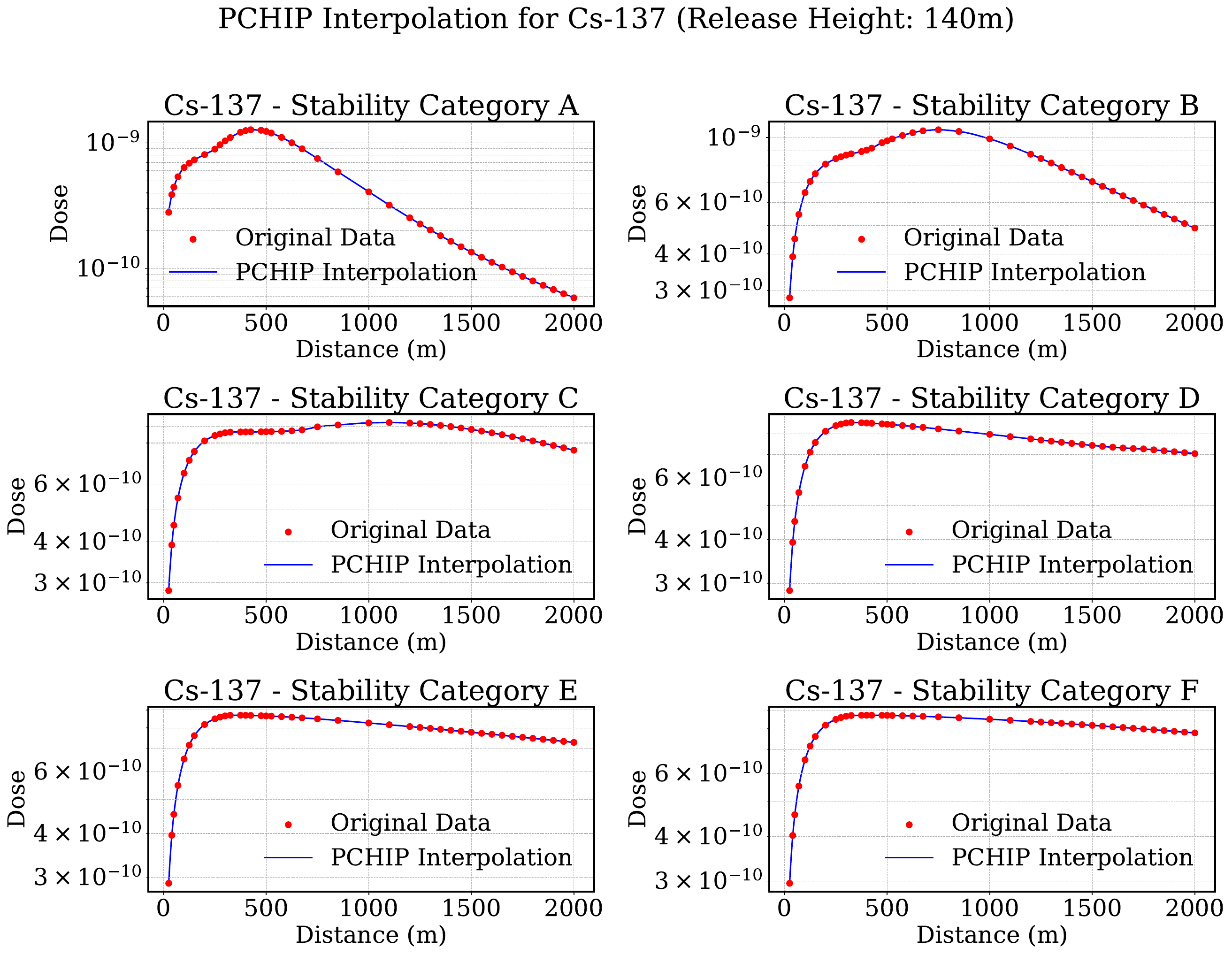}
\caption{Distance-wise validation of PCHIP interpolation for plume shine dose of \ce{^{137}Cs} at a release height of 140 m. The figure compares numerically calculated dose values and PCHIP-interpolated profiles across Pasquill–Gifford stability categories A–F, demonstrating that the interpolation preserves the physically expected monotonic decay with distance without introducing non-physical oscillations.}
\label{fig:1dphcip}
\end{figure}

Using PCHIP interpolation, dose values were generated at uniformly spaced distance points between the minimum and maximum observed distances for each scenario. This procedure enhances spatial resolution while preventing artificial oscillations or non-physical extrema that may arise from higher-order spline methods. Representative examples of the distance-wise interpolation for different radionuclides and stability categories are presented in Figs.~\ref{fig:1dphcip_a}–\ref{fig:1dphcip}.

\subsection{Final Datasets for Model Training and Evaluation}
\label{sec:final_datasets}

The low-resolution $\mathcal{D}_{\mathrm{LR}}$ dataset was partitioned into training ($\mathcal{D}_{\mathrm{LR}}^{\mathrm{train}}$) and test ($\mathcal{D}_{\mathrm{LR}}^{\mathrm{test}}$) subsets, as summarized in Table~\ref{tab:dataset_def}. For learning in the discrete input space, models were trained on $\mathcal{D}_{\mathrm{LR}}^{\mathrm{train}}$ and evaluated on the held-out set $\mathcal{D}_{\mathrm{LR}}^{\mathrm{test}}$. 

The high-resolution $\mathcal{D}_{\mathrm{HR}}$ dataset is exclusively obtained from  $\mathcal{D}_{\mathrm{LR}}^{\mathrm{train}}$ via interpolation. To ensure a purely interpolated domain, all original discrete sampling points present in $\mathcal{D}_{\mathrm{LR}}^{\mathrm{train}}$ were removed, resulting in a refined $\mathcal{D}_{\mathrm{HR}}$. This refined dataset was then randomly partitioned into $\mathcal{D}_{\mathrm{HR}}^{\mathrm{train}}$ (99.975\%) and $\mathcal{D}_{\mathrm{HR}}^{\mathrm{test}}$ (0.025\%), such that both subsets consist solely of interpolated data points and represent a quasi-continuous input domain.

$\mathcal{D}_{\mathrm{LR}}^{\mathrm{test}}$ and $\mathcal{D}_{\mathrm{HR}}^{\mathrm{test}}$ were used to evaluate the performance of models trained on low- and high-resolution data in the original discrete domain and the interpolated domain, respectively. 

This dataset organization enables a two-level evaluation strategy: model accuracy is assessed both at the original numerically computed sampling locations and at previously unseen interpolated points. Together, these tests provide a rigorous measure of predictive fidelity and generalization capability across sparsely sampled regions of the physical parameter space.

\begin{table}[h]
\centering
\caption{Definition of training and test datasets used for model development and evaluation}
\label{tab:dataset_def}
\begin{tabular}{l l p{8.5cm}}
\hline
\textbf{Symbol} & \textbf{Dataset Type} & \textbf{Description and Purpose} \\ \hline
$\mathcal{D}_{\mathrm{LR}}^{\mathrm{train}}$ & Low-resolution training set &
99\% subset of the original discrete dataset ($\sim9\times10^{4}$ samples), used for initial model training on numerically computed plume shine dose values. \\ \hline
$\mathcal{D}_{\mathrm{LR}}^{\mathrm{test}}$ & Low-resolution test set &
1\% held-out subset of the original dataset ($\sim10^{3}$ samples), reserved for final evaluation against unseen numerically computed dose values. \\ \hline
$\mathcal{D}_{\mathrm{HR}}^{\mathrm{train}}$ & High-resolution training set &
Interpolated dataset after removal of all original low-resolution sampling points, comprising approximately 99.975\% of the remaining interpolated samples ($\sim4\times10^{6}$ points), used for training models on a dense and continuous input domain. \\ \hline
$\mathcal{D}_{\mathrm{HR}}^{\mathrm{test}}$ & High-resolution test set &
Held-out subset of the dataset (0.025\%, $\sim10^{3}$ samples), containing only interpolated data points which are used to evaluate model generalization at interpolated input-space. \\ \hline
\end{tabular}
\end{table}

\section{ML Framework and Model Configuration}
\label{sec:models}

To examine how different ML paradigms perform for plume shine dose prediction, three representative regression models were selected: RF, XGBoost, and TabNet. Together, these models cover the main approaches currently used for structured tabular data, namely ensemble learning, boosted decision trees, and attention-based DL model. 

The choice of these models is motivated by two considerations central to this study. First, plume shine dose prediction is governed by a small set of physically meaningful variables, including geometric features such as downwind distance and release height, as well as atmospheric dispersion conditions. These variables directly define the source–receptor geometry and therefore play a dominant role in dose formation. This structure makes plume shine dose prediction a suitable test case for examining how different model inductive biases learn and prioritize geometry-driven relationships, which are analyzed in detail in the Results section.. Second, while tree-based ensemble methods are widely regarded as strong performers on tabular data, recent DL architectures such as TabNet aim to challenge this dominance by introducing attention mechanisms tailored to structured inputs. Comparing these approaches therefore provides insight into which modeling strategies are best aligned with the physics and data structure of radiological dose assessment. 

Below, we provide a brief overview of these models as an introduction. 

\subsection{Overview of ML Models}
\label{sec:ml_models}

RF constructs multiple decision trees using bootstrapped samples of the training data and aggregates their predictions through averaging \cite{Breiman2001rf}. Random feature selection at each split reduces correlation among trees and improves generalization. RF regression was implemented using the \texttt{scikit-learn} framework \cite{pedregosa2011scikit}.

XGBoost is a gradient boosting approach in which decision trees are constructed sequentially, with each new tree trained to correct the residual errors of the existing ensemble \cite{chen2016xgboost}. The framework includes regularization and structural constraints that help limit model complexity and improve predictive stability. In this work, XGBoost regression was implemented using the official \texttt{xgboost} library.

TabNet is a DL architecture developed specifically for learning from tabular data \cite{arik2021tabnet}. It uses a sequential attention mechanism to perform instance-wise feature selection at each decision step, followed by feature transformation layers whose contributions are combined to generate the final prediction. In this study, TabNet was implemented using the official PyTorch-based implementation\cite{paszke2019pytorch}.

Details of model hyperparameter selection and tuning procedures are discussed in the following subsection.

\subsection{Data Preprocessing}
\label{sec:data_preprocess}

Prior to model training, all datasets were preprocessed to ensure numerical stability and compatibility across the three ML architectures. The preprocessing workflow comprised target variable transformation, categorical feature encoding, and normalization of numerical features.

\begin{figure}[!htbp]
    \centering
    \includegraphics[width=0.9\linewidth]{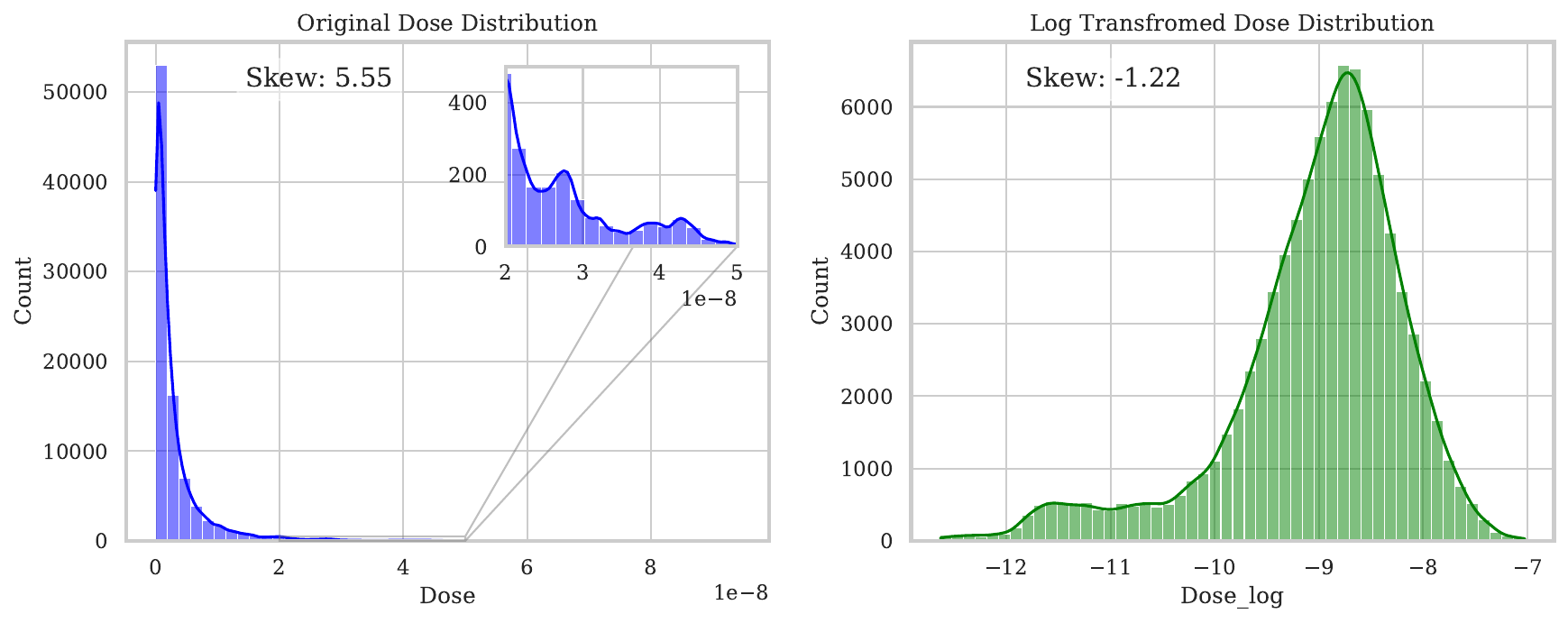}
    \caption{Histogram comparison of plume shine dose distributions before and after base-10 logarithmic transformation.}
    \label{fig:Comp_histplot}
\end{figure}

The plume shine dose, used as the target variable, exhibits a strongly right-skewed distribution spanning several orders of magnitude. A base-10 logarithmic scaling was applied to the dose values to reduce the dominance of extreme values and yield a more symmetric distribution that is better suited for regression-based learning. As illustrated in Fig.~\ref{fig:Comp_histplot}, this transformation substantially reduces skewness (from 5.55 to $-1.22$), promoting stable optimization and improved convergence. Model predictions were inverse-transformed during post-processing to recover dose values in physical units for evaluation and interpretation.

Categorical features, namely \textit{Radionuclide} and \textit{Atmospheric Stability Category}, were encoded using integer representations to preserve category identity while maintaining consistency across models. For TabNet, these encoded features were explicitly specified through the \texttt{cat\_idxs} and \texttt{cat\_dims} parameters, which define the indices and cardinalities of categorical variables, respectively. Numerical features, including \textit{Release Height} and \textit{Downwind Distance}, were normalized to the range $[0,1]$ using Min--Max scaling to prevent scale imbalance and ensure stable model training.

\subsection{Hyperparameter Optimization}
\label{sec:hyperparam_opt}

Hyperparameter optimization was performed separately for each model, accounting for differences in model complexity and computational cost. For RF and XGBoost, which are comparatively less computationally intensive, hyperparameters were tuned using a manual, iterative procedure. Key parameters were adjusted heuristically based on observed training and validation performance, with the objective of achieving a balanced trade-off between bias and variance.

For TabNet, a hybrid optimization strategy was adopted due to its higher sensitivity to hyperparameter selection and increased computational demands. An initial manual search was conducted to identify reasonable ranges for critical hyperparameters. Subsequently, automated hyperparameter optimization was performed using the Optuna framework \citep{akiba2019optuna}. Fifty optimization trials were executed using Optuna’s Tree-structured Parzen Estimator (TPE) sampler, with the objective of minimizing the Root Mean Square Error (RMSE) on the validation set. The tables with optimized hyperparaters are provided in the supporting information (Table S1, S2, S3)


\subsection{Performance Metrics for Evaluation and Testing}

During training, the optimization criteria differed across models. XGBoost and TabNet were trained by minimizing the Root Mean Square Error (RMSE), while the RF model employed Mean Squared Error (MSE), which is its default loss function. These training losses were used solely for parameter optimization and are not treated as primary comparative metrics in the evaluation on unseen test data.

To assess the predictive performance of each model, evaluation metrics that capture both goodness of fit and relative prediction accuracy were employed. Specifically, three complementary metrics were used: the coefficient of determination ($R^2$), Mean Absolute Percentage Error (MAPE), and symmetric Mean Absolute Percentage Error (sMAPE).

$R^2$ quantifies the proportion of variance in the reference plume shine dose explained by the model predictions and is defined as
\begin{equation}
R^2 = 1 - \frac{\sum_{i=1}^{N} (y_i - \hat{y}_i)^2}{\sum_{i=1}^{N} (y_i - \bar{y})^2},
\end{equation}
where $y_i$ and $\hat{y}_i$ denote the true and predicted dose values, respectively, $\bar{y}$ is the mean of the true values, and $N$ is the number of test samples.

MAPE measures the average relative deviation between predicted and reference doses and is given by
\begin{equation}
\mathrm{MAPE} = \frac{100}{N} \sum_{i=1}^{N} 
\left| \frac{y_i - \hat{y}_i}{y_i} \right|.
\end{equation}
MAPE provides an intuitive percentage-based measure of error; however, it can become unstable or unbounded when reference dose values are very small.

To mitigate this limitation, sMAPE is also employed, which normalizes the absolute error by the average magnitude of the true and predicted values and is defined as
\begin{equation}
\mathrm{sMAPE} = \frac{100}{N} \sum_{i=1}^{N} 
\frac{2 \left| y_i - \hat{y}_i \right|}{\left| y_i \right| + \left| \hat{y}_i \right|}.
\end{equation}

Unlike MAPE, which can theoretically range from $0$ to $\infty$, sMAPE is bounded within the interval $[0, 200]$ and treats overestimation and underestimation symmetrically. This property is particularly important in the present study, as plume shine dose values often span several orders of magnitude and include extremely small values. Under such conditions, absolute error metrics such as RMSE or MAE may become less informative for comparative evaluation, whereas percentage-based metrics provide a more meaningful assessment of model performance across the full dose range.

\section{Results and Discussion}

\subsection{Model Performance under Discrete and Continuous Data Regimes}
\label{sec:results_models}

This subsection examines how training data resolution influences predictive accuracy and generalization behavior across the three models. Performance is evaluated in two complementary settings: learning from discrete low-resolution dose tables and learning from interpolation-enhanced high-resolution dense datasets.

\paragraph{Case I: Training on Low-Resolution Data.}

As summarized in Table~\ref{tab:model_perf_og}, both XGBoost and RF showed strong predictive performance on $\mathcal{D}_{\mathrm{LR}}^{\mathrm{test}}$ when trained exclusively on $\mathcal{D}_{\mathrm{LR}}^{\mathrm{train}}$ (Table \ref{tab:dataset_def}). XGBoost yielded the highest coefficient of determination ($R^2 = 0.999$) along with very low error levels (MAPE and sMAPE below 0.5\%), indicating excellent agreement with the numerically computed reference dose values at discrete sampling locations. RF also performed well, achieving similar $R^2$ of $\approx$ 0.99 but with slightly higher MAPE and sMAPE error values around 2\%. In contrast, TabNet exhibited noticeably weaker performance, with a lower $R^2$ value of 0.9561 and substantially higher percentage-based errors. This degradation possibly reflects the challenges faced by attention-based neural networks when trained on sparse and discontinuous datasets in which limited sampling density hinders the learning of stable feature representations. Figure S1 (see Supporting information) provides a visual comparison of predicted versus reference plume shine dose values and confirms the quantitative trends reported in Table~\ref{tab:model_perf_og}. The tight clustering around the one-to-one line for XGBoost and RF contrasts with the broader scatter observed for TabNet, consistent with the corresponding differences in $R^2$ and error metrics.

\paragraph{Poor Generalization to High-Resolution Test Data.}

When the models trained on low-resolution dataset (i.e. $\mathcal{D}_{\mathrm{LR}}^{\mathrm{train}}$) were further evaluated on interpolated test dataset ($\mathcal{D}_{\mathrm{HR}}^{\mathrm{test}}$), which consists exclusively of interpolated points, a clear reduction in predictive performance was observed across all three models. As summarized in Table~\ref{tab:model_perf_og}, the error metrics increased substantially compared to the evaluation in $\mathcal{D}_{\mathrm{LR}}^{\mathrm{test}}$, indicating limited generalization beyond the discrete training grid. Although XGBoost retained a high coefficient of determination ($R^2 = 0.999$), its percentage-based errors increased noticeably, with MAPE rising to 2.47\% and sMAPE to 2.42\%. A similar trend was observed for RF, where MAPE and sMAPE increased to 3.16\% and 3.15\%, respectively. These results also indicate that $R^2$ is alone insufficient to characterize generalization quality in dose prediction, as small relative deviations can translate into meaningful absolute errors over several orders of magnitude.

TabNet exhibited the most pronounced degradation, with $R^2$ decreasing to 0.9157 and error metrics exceeding 14\% (Fig. S2). This behavior again reiterates the sensitivity of attention-based TabNet to sparse training distributions, particularly when predictions are required at intermediate locations not explicitly represented in the training data. 

Overall, these results demonstrated that models trained solely on $\mathcal{D}_{\mathrm{LR}}$ primarily learn to reproduce discrete dose tables and do not reliably capture smooth dose variation across the continuous physical domain. The pronounced performance degradation observed when predictions are evaluated at interpolated locations underscores the limited generalization capability of learning from sparsely sampled data. Therefore, these findings advocate for the training of models with high-resolution training data set.

\paragraph{Case II: Training on High-Resolution Data.}

Motivated by the limited generalization observed in Case I, all three models were retrained using the high-resolution dataset ($\mathcal{D}_{\mathrm{HR}}^{\mathrm{train}}$), which provides dense sampling along the dimensions of downwind distance. The retrained models were then evaluated on both the ($\mathcal{D}_{\mathrm{LR}}^{\mathrm{test}}$) and $\mathcal{D}_{\mathrm{HR}}^{\mathrm{test}}$. This configuration enables assessment of (i) backward consistency with physics-based discrete calculations and (ii) generalization across previously unseen continuous regions of the input space.

Figure S3 (see Supporting information) showed that models trained on $\mathcal{D}_{\mathrm{HR}}^{\mathrm{train}}$ retain excellent predictive accuracy when evaluated on the low-resolution test data ($\mathcal{D}_{\mathrm{LR}}^{\mathrm{test}}$). This indicates that training on interpolated data does not degrade fidelity to the original physics-based reference points. Notably, XGBoost achieves an $R^2$ of 0.9996 with less than 1\% MAPE, while RF exhibits comparable performance with slightly higher percentage errors.  Further, evaluation on the high-resolution test set (Fig. S4) also demonstrates a substantial improvement in generalization compared to Case I. All models show reduced errors and improved alignment with reference dose values across the interpolated domain. As summarized in Table~\ref{tab:model_perf_og}, XGBoost maintains the strongest overall performance, with approximately 1\% MAPE on $\mathcal{D}_{\mathrm{HR}}^{\mathrm{test}}$, confirming its ability to robustly capture the underlying functional relationship governing plume shine dose. RF shows moderate degradation relative to XGBoost but the model performed significantly better than the trained model with $\mathcal{D}_{\mathrm{LR}}^{\mathrm{train}}$.


\begin{table}[!htbp]
\centering
\caption{Performance of ML models on plume shine dose predictions evaluated on the original (physical) dose scale}
\label{tab:model_perf_og}
\renewcommand{\arraystretch}{1.2}

\begin{tabular}{p{0.23\textwidth} c c c c c c c c c}
\hline
\multirow{2}{*}{\textbf{Training--Testing}} &
\multicolumn{3}{c}{\textbf{XGBoost}} &
\multicolumn{3}{c}{\textbf{RF}} &
\multicolumn{3}{c}{\textbf{TabNet}} \\ 

& $R^2$ & MAPE & sMAPE
& $R^2$ & MAPE & sMAPE
& $R^2$ & MAPE & sMAPE \\ \hline

$\mathcal{D}_{\mathrm{LR}}^{\mathrm{train}} \rightarrow \mathcal{D}_{\mathrm{LR}}^{\mathrm{test}}$
& \textbf{0.999} & \textbf{0.48} & \textbf{0.48}
& 0.9979 & 1.92 & 1.91
& 0.9561 & 13.90 & 13.94 \\

$\mathcal{D}_{\mathrm{LR}}^{\mathrm{train}} \rightarrow \mathcal{D}_{\mathrm{HR}}^{\mathrm{test}}$
& \textbf{0.999} & \textbf{2.47} & \textbf{2.42}
& 0.9956 & 3.16 & 3.15
& 0.9157 & 14.41 & 13.97 \\

$\mathcal{D}_{\mathrm{HR}}^{\mathrm{train}} \rightarrow \mathcal{D}_{\mathrm{LR}}^{\mathrm{test}}$
& \textbf{0.999} & \textbf{0.93} & \textbf{0.93}
& 0.9975 & 2.07 & 2.06
& 0.9983 & 3.97 & 3.88 \\

$\mathcal{D}_{\mathrm{HR}}^{\mathrm{train}} \rightarrow \mathcal{D}_{\mathrm{HR}}^{\mathrm{test}}$
& \textbf{0.998} & \textbf{1.00} & \textbf{1.00}
& 0.9930 & 2.22 & 2.22
& 0.9833 & 3.43 & 3.36 \\

\hline
\end{tabular}
\end{table}

 Interestingly, trained TabNet model showed a marked improvement relative to Case I for both $\mathcal{D}_{\mathrm{LR}}^{\mathrm{test}}$ and $\mathcal{D}_{\mathrm{HR}}^{\mathrm{test}}$ dataset, achieving an $R^2$ of $\approx$ 0.99 with significantly reduced percentage-based errors. This improvement highlights the benefit of dense training data for DL architectures\cite{ye2024closer}, which typically require smooth and continuous input–output mappings to extract stable feature representations.

Taken together, these findings established the high-resolution interpolated dataset as a necessary component for reliable surrogate modeling of plume shine dose and justify the interpolation-assisted learning strategy adopted in this work. As discussed, models trained solely on low-resolution data can accurately reproduce known discrete points, but they fail to infer smooth transitions between them. This behavior is consistent with recent findings\cite{grinsztajn2022tabular} on tabular ML approaches, where the inductive biases of different model families lead to markedly different generalization behavior on irregular and sparsely sampled targets like those in plume shine dose tables. The high-resolution training dataset mitigates this limitation by encoding physically consistent trends across distance, enabling all models --- particularly XGBoost and TabNet --- to generalize effectively to both discrete numerically computed locations and unseen interpolated regions.

\subsection{Model Robustness Across Atmospheric Regimes and Radionuclide Classes}

\begin{figure}[!htbp]
    \centering
    \includegraphics[width=\textwidth]{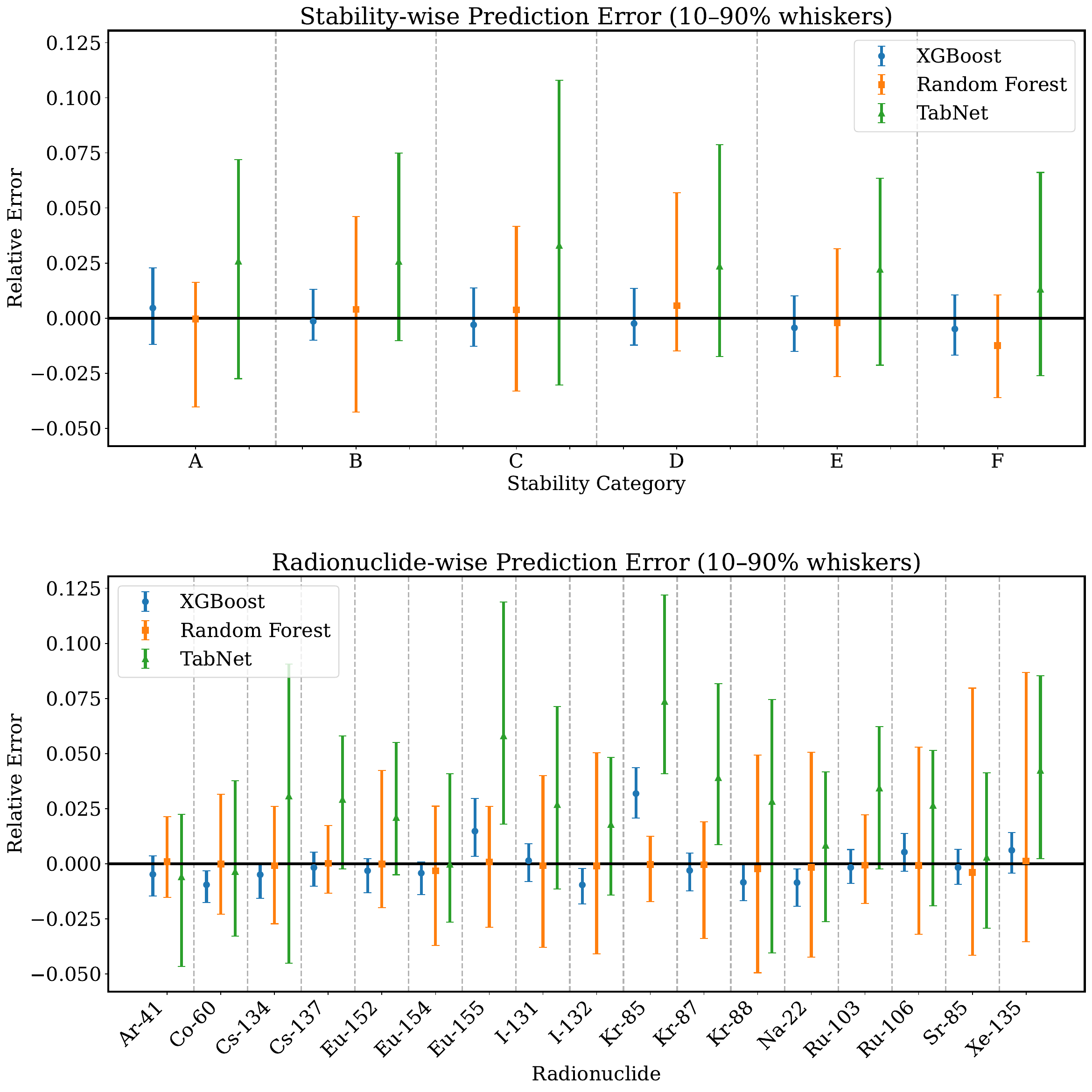}
    \caption{Stability-wise (top) and radionuclide-wise (bottom) relative error distributions for XGBoost, RF, and TabNet. Markers denote median relative errors, while whiskers represent the 10$^{th}$--90$^{th}$ percentile range, computed over the combined interpolated and non-interpolated test datasets.}
    \label{fig:whisker_plot}
\end{figure}

To gain a deeper understanding of model performance at the regime level for the interpolated dataset, relative error distributions were analyzed using whisker plots (Fig.~\ref{fig:whisker_plot}), separately for each atmospheric stability class and radionuclide, for XGBoost, RF, and TabNet. Across all stability categories (Fig.~\ref{fig:whisker_plot}, top), XGBoost exhibited the smallest median relative errors and the narrowest 10–90\% uncertainty bounds, indicating consistently stable performance with minimal systematic bias. Although RF maintained good central (median) accuracy, its variance found to be on the higher side. TabNet displayed the largest spread of errors across all the stability classes, with most pronounced upper whiskers for stability category (Fig.~\ref{fig:whisker_plot}, top).

The radionuclide-wise error distributions in Fig.~\ref{fig:whisker_plot} (bottom) indicated that predictive performance is primarily influenced by the intrinsic structure and smoothness of the plume-shine dose field associated with each isotope. No strict monotonic trend was observed with respect to gamma emission strength or half-life alone, which is expected since, under short-term steady plume release conditions, radioactive decay during atmospheric transport is negligible compared to dispersion-driven effects. Instead, radionuclides with softer gamma spectra (e.g., Eu isotopes) and non-depositing noble gases exhibit comparatively larger error spread, attributable to stronger atmospheric attenuation and steeper dose–distance gradients that introduce higher functional nonlinearity. In contrast, high-energy gamma emitters (e.g., \ce{^{22}Na} and \ce{^{41}Ar}), as well as longer-lived radionuclides such as \ce{^{137}Cs} and \ce{^{60}Co}, display comparatively smoother spatial attenuation and more regular dose–distance relationships, leading to more stable prediction performance. Larger uncertainties were particularly evident for noble gases such as \ce{^{85}Kr}, \ce{^{87}Kr}, and \ce{^{135}Xe}, especially for RF and TabNet. Owing to their chemically inert and non-depositing nature, noble gas plumes lack environmental removal mechanisms such as dry and wet deposition, resulting in highly localized centerline concentrations and strong spatial dose gradients, particularly under stable atmospheric conditions. Consequently, the observed radionuclide-dependent variability in the error statistics is governed more by the smoothness and curvature of the underlying dose field—controlled by photon energy and dispersion characteristics—than by radionuclide half-life or identity alone.

\begin{figure}[!htbp]
    \centering
    \includegraphics[width=\textwidth]{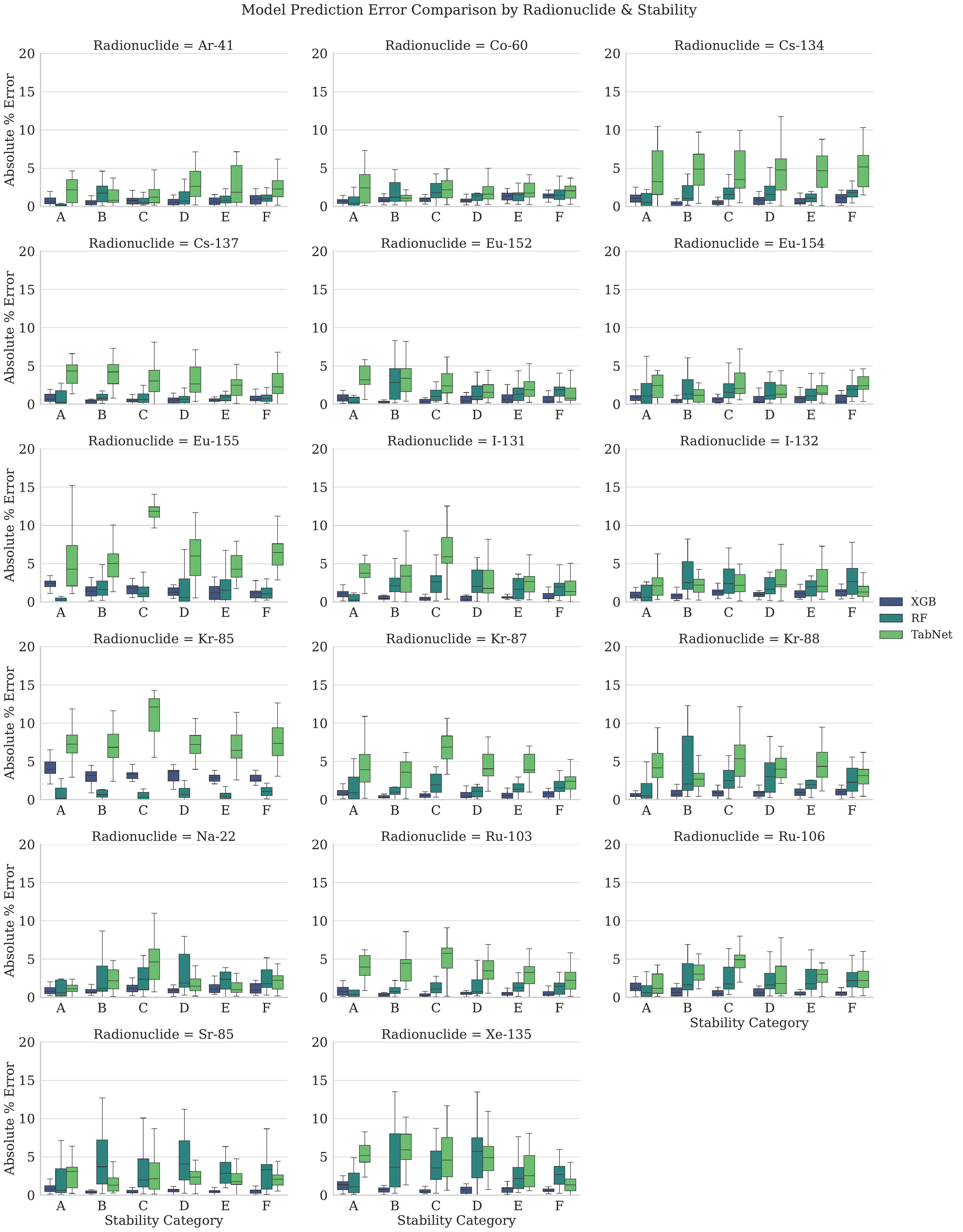}
    \caption{Faceted distributions of absolute percentage error by radionuclide and atmospheric stability category, highlighting regime-dependent variations in model performance for XGBoost, RF, and TabNet.}
    \label{fig:faceted_plot}
\end{figure}

Further, the faceted error distributions in Fig.~\ref{fig:faceted_plot} provide a detailed diagnostic view of prediction accuracy across radionuclide type and atmospheric stability categories. The plots reveal clear regime-dependent behavior, with RF and TabNet generally showing larger error spread and more frequent outliers across several stability classes, particularly for noble gas radionuclides, whereas XGBoost maintains comparatively tighter and more consistent error distributions, reiterating the earlier radionuclide-wise observation (Fig. \ref{fig:whisker_plot}).  

Undoubtedly, the consistently lower bias and variance exhibited by XGBoost across both stability-wise and radionuclide-wise analyses underscore its suitability as a \textit{primary surrogate model for plume-shine dose estimation}. In contrast, the comparatively higher variability observed for RF and TabNet suggests that caution is warranted when deploying these models under regimes with strong spatial dose gradients, particularly in stability-dependent dispersion conditions.

\subsection{Radionuclide-Wise Feature Importance and Model Interpretation}
\subsubsection{Interpretability metrics and evaluation protocol.}
To ensure a consistent and model-agnostic comparison of feature relevance, all interpretability analyses were conducted on a unified test dataset defined as
\begin{equation}
\mathcal{D}_{\mathrm{test}} = 
\mathcal{D}_{\mathrm{LR}}^{\mathrm{test}} \cup 
\mathcal{D}_{\mathrm{HR}}^{\mathrm{test}} .
\end{equation}

For the tree-based models, feature relevance was quantified using \emph{conditional permutation importance}. Specifically, for a given radionuclide $r$ and feature $f$, the importance was computed as the expected increase in prediction error obtained by randomly permuting feature $f$ within the radionuclide-specific subset of the test data,
\begin{equation}
\mathcal{I}_{r}(f)
=
\mathbb{E}_{(x,y)\sim \mathcal{D}_{\mathrm{test}}(r)}
\left[
\mathcal{L}\big(y, \hat{y}(x_{f}^{\pi})\big)
-
\mathcal{L}\big(y, \hat{y}(x)\big)
\right],
\end{equation}

where $\mathcal{D}_{\mathrm{test}}(r)$ denotes the subset of test samples corresponding to radionuclide $r$, $x_{f}^{\pi}$ represents the input vector in which feature $f$ has been permuted conditional on the remaining features, and $\mathcal{L}(\cdot)$ denotes the mean squared error.

For TabNet, interpretability was obtained from the model’s intrinsic attention mechanism. 
At each decision step $t$, TabNet computes a feature mask $M_t^{(i)}(f)$ for sample $i$, 
which represents the relative contribution of feature $f$ at that step.

Radionuclide-wise feature importance was then computed by aggregating these masks across all samples 
and decision steps in the unified test dataset,
\begin{equation}
\mathcal{A}_{r}(f)
=
\frac{1}{\lvert \mathcal{D}_{\mathrm{test}}(r) \rvert}
\sum_{i \in \mathcal{D}_{\mathrm{test}}(r)}
\sum_{t=1}^{T} M_t^{(i)}(f),
\end{equation}
where $\mathcal{D}_{\mathrm{test}}(r)$ denotes the subset of test samples corresponding to radionuclide $r$, 
and $T$ is the total number of TabNet decision steps.

This formulation yields a population-level measure of feature utilization that reflects the internal 
reasoning strategy learned by the TabNet model.

For visualization and cross-model comparison, all radionuclide-wise importance vectors were row-wise normalized so that the feature importances for each radionuclide sum to unity. This transformation yields relative importance profiles that highlight feature dominance patterns within each radionuclide while enabling consistent qualitative comparison across RF, XGBoost, and TabNet, despite the differing attribution mechanisms (permutation-based vs. attention-based).

\begin{figure}[!htbp]
    \centering
    \includegraphics[width=1\linewidth]{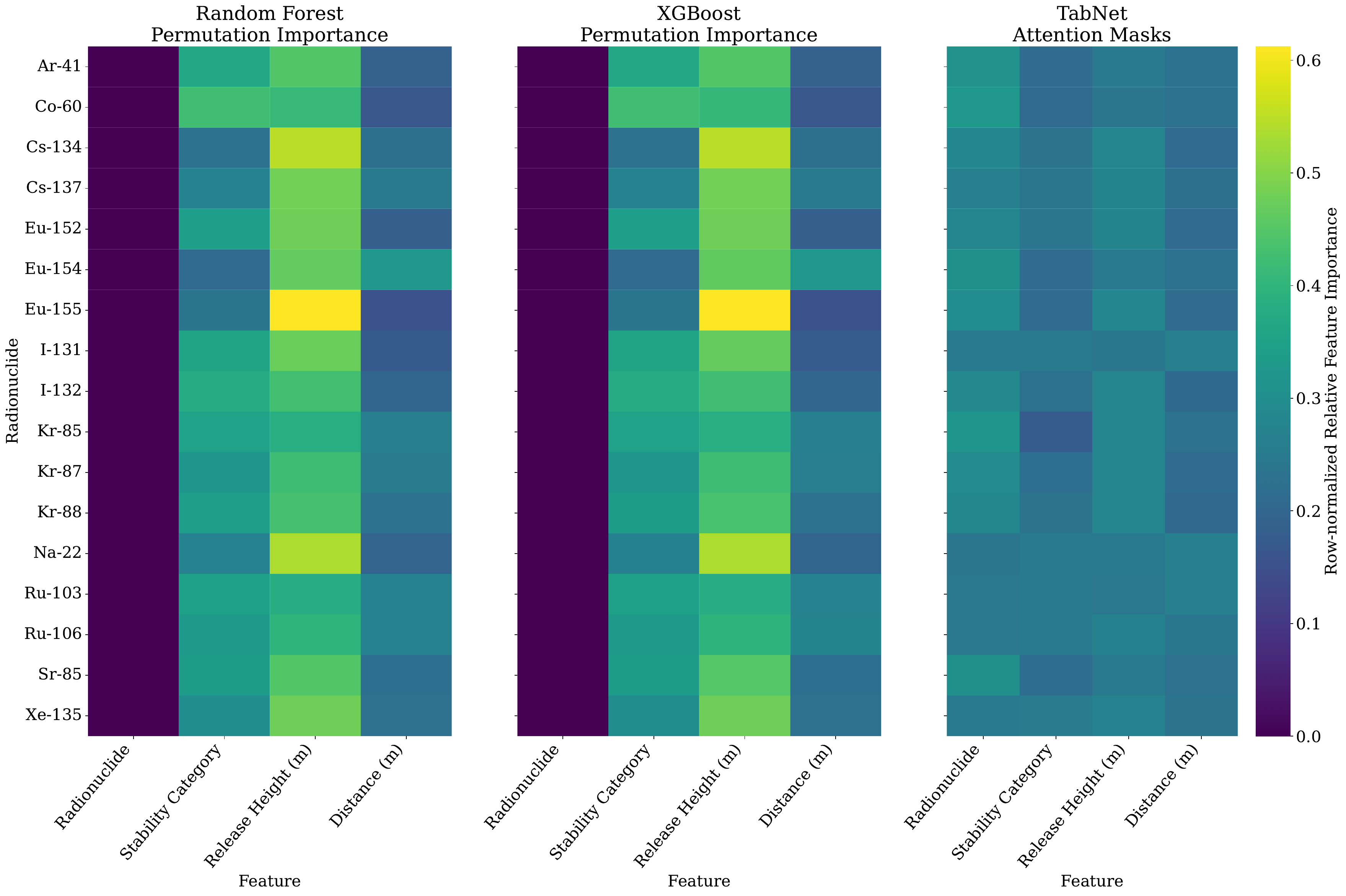}
    \caption{Radionuclide-wise feature importance across RF, XGBoost, and TabNet model trained on $\mathcal{D}_{\mathrm{HR}}^{\mathrm{train}}$. RF and XGBoost importance is computed via radionuclide-conditional permutation importance, while TabNet importance is obtained from aggregated attention masks. Importance values are row-wise normalized for each radionuclide, allowing relative comparison of feature dominance patterns across models under a shared color scale, rather than absolute magnitude equivalence between permutation- and attention-based attributions.}

    \label{fig:feature_imp_3panel}
\end{figure}

\subsubsection{Qualitative structure of the comparative feature-importance maps}

The comparative feature-importance maps in Fig.~\ref{fig:feature_imp_3panel} highlight both shared physical trends and model-specific differences in how input features are utilized. Since the importance values are row-wise normalized for each radionuclide, the maps should be interpreted as showing the relative contribution of features within each radionuclide under a common color scale, rather than absolute importance magnitudes.

For both XGBoost and RF, release height consistently emerges as the dominant contributor to plume shine dose prediction across nearly all radionuclides, with atmospheric stability and downwind distance acting as a secondary controls. The near-identical importance patterns observed for the two tree-based models indicate that both capture the same underlying geometric and dispersion-driven dependencies governing the dose field. Such sharply concentrated, dominance-oriented feature hierarchies are commonly observed in tree-based models, as they tend to prioritize features that provide the largest error reduction through hierarchical splits. In this context, the consistently negligible importance assigned to radionuclide identity further suggests that predictions are primarily controlled by plume geometry and associated dispersion related conditions rather than isotope-specific characteristics. 

In contrast, the feature-importance patterns from TabNet appear distinctly different in which the importance values are more evenly spread across all features, with no single variable exhibiting strong dominance. This smoother allocation suggests that TabNet relies on a more balanced combination of inputs rather than sharply prioritizing specific physical controls. 

\subsubsection{Geometry-dispersion-only versus full-feature ablation.}

\begin{figure}[!htbp]
    \centering
    \begin{subfigure}[t]{0.49\linewidth}
        \centering
        \includegraphics[width=\linewidth]{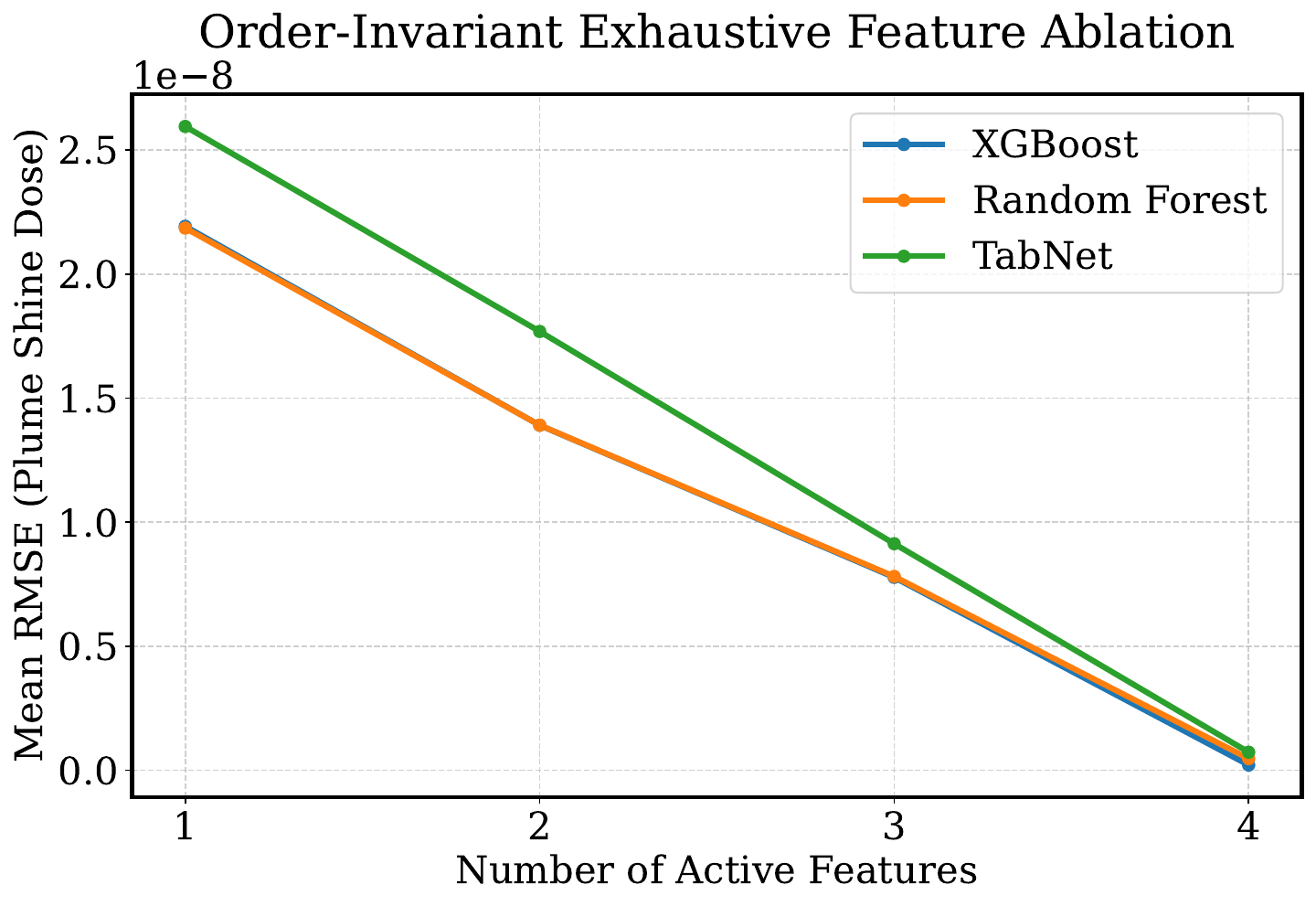}
        \caption{}
    \end{subfigure}
    \hfill
    \begin{subfigure}[t]{0.49\linewidth}
        \centering
        \includegraphics[width=\linewidth]{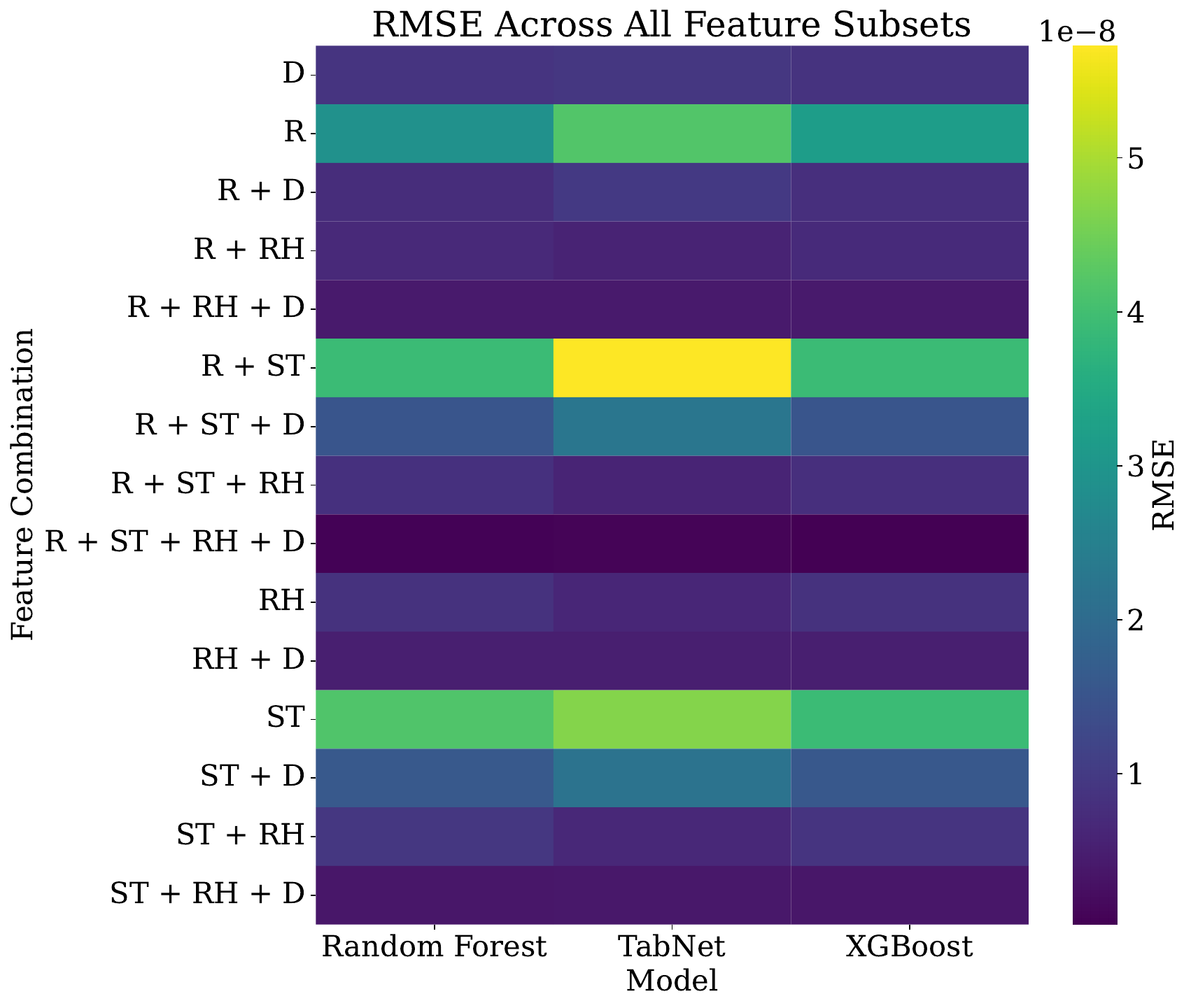}
        \caption{}
    \end{subfigure}
    
    \caption{Ablation analysis for plume-shine dose prediction: (a) Order-invariant mean ablation curve showing mean RMSE (physical dose space) versus number of active features, averaged over all possible feature subsets. (b) The heatmap shows that feature combinations containing release height, stability category, and distance consistently yield lower prediction errors across RF, XGBoost, and TabNet, while the addition of radionuclide identity provides comparatively smaller marginal improvement. Acronyms: R = Radionuclide, ST = Stability Category, RH = Release Height, and D = Downwind Distance.}
    \label{fig:exhaustive_ablation_combined}
\end{figure}

An order-invariant exhaustive ablation study was conducted to systematically assess the contribution of each input feature to plume-shine dose prediction by evaluating model performance over all possible feature combinations, thereby eliminating any bias associated with feature ordering. For each subset, the mean RMSE was computed on the predictions, followed by aggregation across subsets with the same number of active features. As shown in the exhaustive ablation curve (Fig. \ref{fig:exhaustive_ablation_combined}a), all three models exhibit a monotonic decrease in RMSE as additional features are introduced; however, the most substantial error reduction occurs when increasing the feature set from one to three variables, while the marginal improvement from the fourth feature is comparatively small. This pattern reiterated that the dominant predictive structure is captured primarily by the geometry–dispersion features (release height, stability category, and downwind distance), with radionuclide identity contributing mainly as a secondary conditional refinement. 

The subset heatmap (Fig. \ref{fig:exhaustive_ablation_combined}b) further corroborates this observation by showing consistently lower RMSE for combinations that include release height and stability, whereas subsets lacking these variables yield significantly higher errors across all models. Notably, XGBoost achieves the lowest RMSE across nearly all subset configurations, followed closely by Random Forest, while TabNet consistently exhibits higher error, suggesting that tree-based ensembles more efficiently exploit the low-dimensional geometry–dispersion structure governing the dose field. 


\subsubsection{Origin of the performance hierarchy among models}

The observed performance ordering, XGBoost $>$ RF $>$ TabNet, can be understood in terms of the interaction between model inductive bias and the underlying physical structure of the plume-shine dose field, which is primarily governed by geometry–dispersion features (release height, stability category, and downwind distance), with radionuclide identity acting as a secondary conditional modifier. The exhaustive ablation analysis further supports this interpretation by showing that the majority of error reduction is achieved when geometry–dispersion variables are included, while the marginal gain from radionuclide information remains comparatively smaller.

XGBoost is particularly well aligned with this structure. Its sequential boosting mechanism effectively captures the dominant geometry–dispersion dependencies and incrementally refines residual errors associated with conditional effects such as radionuclide characteristics and stability-dependent dispersion. This residual-correction behavior is consistent with prior findings that boosted tree ensembles perform strongly on tabular problems governed by low-dimensional and smooth functional relationships \cite{grinsztajn2022tabular,shwartz2022tabular}, thereby explaining its consistently superior predictive accuracy.

RF, in contrast, captures the dominant geometry–dispersion trends in a robust and stable manner but exhibits slightly reduced sensitivity to finer conditional refinements. This behavior is characteristic of bagged tree ensembles, which emphasize stable global patterns through averaging rather than sequential residual correction \cite{breiman2001random}. As a result, while RF achieves performance close to XGBoost, it derives comparatively smaller gains from additional conditional information, consistent with the ablation results.

TabNet exhibits a different inductive bias due to its instance-wise attention mechanism, which distributes feature utilization more flexibly across samples instead of enforcing a single global importance hierarchy. Although this leads to richer and more heterogeneous feature usage, it can result in less efficient allocation of model capacity in settings where a low-dimensional geometry–dispersion structure dominates the target function. This observation is consistent with recent studies showing that deep learning models for tabular data often underperform tree-based ensembles when strong low-order structure governs the prediction task \cite{grinsztajn2022tabular,shwartz2022tabular}. 

Overall, the performance hierarchy therefore arises not merely from differences in model expressiveness, but from how effectively each architecture concentrates its learning on the dominant geometry–dispersion controls of the plume-shine dose field while incorporating radionuclide-dependent effects as secondary refinements.

\begin{figure}[!htbp]
    \centering
    \includegraphics[width=1\linewidth]{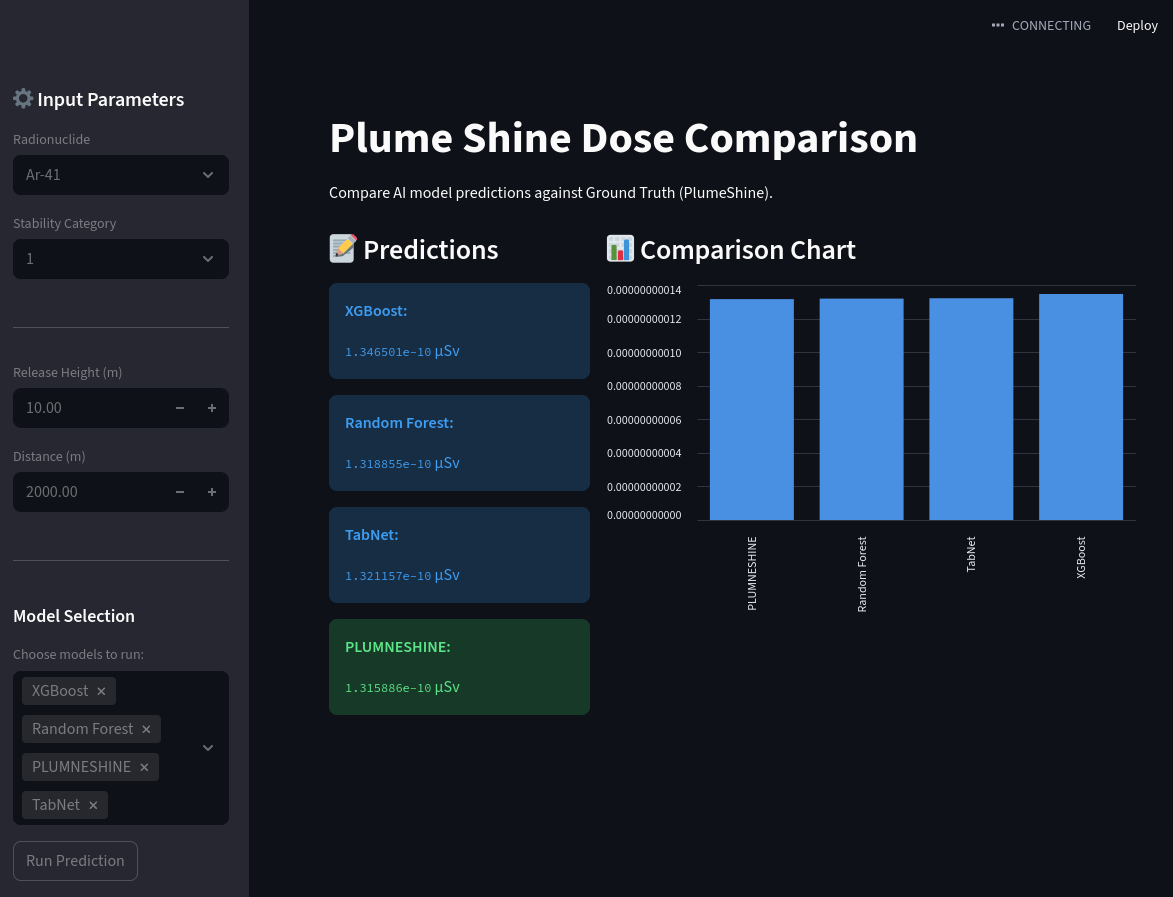}
    \caption{Web-based graphical user interface for interactive plume-shine dose assessment, illustrating scenario input parameters (radionuclide, downwind distance, release height, and atmospheric stability) and the corresponding dose predictions obtained from photon-transport-based calculations (labeled as PLUMEDOSENET in GUI interface) and trained machine-learning models.}
    \label{fig:gui_pdosenet}
\end{figure}

\subsection{Web-based deployment and model accessibility.}

To enhance the practical usability of the proposed surrogate modeling framework, a web-based graphical user interface (GUI) has been developed for interactive plume shine dose assessment. The interface allows users to specify key scenario parameters, namely radionuclide type, downwind distance, release height, and atmospheric stability category. Based on these inputs, plume shine dose predictions are generated independently using the trained XGBoost, RF, and TabNet models. In addition, the GUI computes the corresponding reference dose values using adaptive Gauss quadrature method following Eq. \ref{ps_eq}, enabling direct side-by-side comparison between numerical and machine-learning-based predictions. This deployment-oriented implementation facilitates rapid scenario evaluation, transparent model inter-comparison, and user-driven sensitivity exploration, thereby bridging the gap between methodological development and operational radiological consequence assessment.

\section{Conclusion}

The application of ML to radiation dose assessment remains nontrivial due to safety-critical requirements, limited training-ready datasets, and the strong role of governing physical processes. Plume shine dose estimation provides a relevant benchmark problem in this regard, as it is essential for safety analysis and emergency response, yet remains computationally demanding when evaluated using photon-transport-based approaches. To address these intriguing challenges, this work developed an interpolation-assisted machine learning framework based on discrete plume-shine dose datasets generated using the pyDOSEIA program suite for 17 gamma-emitting radionuclides across multiple dispersion scenarios. These datasets were subsequently augmented using shape-preserving interpolation method to construct dense, high-resolution training datasets suitable for ML-based modeling. Two tree-based ML models--XGBoost and RF--and one deep neural network–based model, TabNet, were systematically evaluated for the plume shine dose prediction. 

The results demonstrate that the interpolation-assisted data augmentation strategy substantially improves generalization across distances, release heights, and radionuclide classes. Among all the ML models, tree-based ensemble methods--particularly XGBoost--achieve the highest prediction accuracy. RF provides robust performance but is less sensitive to finer variations, while TabNet does not yield comparable accuracy gains for this problem.

Further, interpretability analysis using permutation importance (for tree-based models) and attention-based feature attribution (for TabNet) provided useful insight into the factors governing model behavior. The results indicate that performance differences are largely linked to how each model utilizes input features, with TabNet distributing attention across a broader set of variables, while tree-based models concentrate more strongly on dominant geometry–dispersion features, treating radionuclide identity primarily as a secondary conditional input. This difference in feature utilization is consistent with the ablation findings and helps explain the observed variation in prediction accuracy across model architectures.

To support practical application of the proposed framework, a web-based graphical user interface was developed, allowing interactive scenario evaluation and direct comparison between machine-learning-based predictions and photon-transport-based reference calculations. Together, these elements enhance both the transparency and usability of the proposed approach for plume shine dose assessment. Overall, the proposed interpolation-assisted ML framework provides a fast and accurate surrogate approach for plume shine dose estimation, offering a practical alternative to conventional numerical calculations.

\bmhead{Supplementary information}

Detailed information on the optimized hyperparameters, along with additional performance plots obtained from model evaluations on datasets of varying resolutions, is provided in the Supplementary Information.

\bmhead{Acknowledgements}

B.S. and S.A. extend their thanks to Shri Probal Chaudhury (Associate Director, HS \& E Group) for his unwavering support and encouragement throughout the project.

\section*{Declarations}

\bmhead{Code availability}

The source code of this work is accessible at: https://github.com/BiswajitSadhu/PlumeDoseNet. The dataset and trained models are available at https://doi.org/10.5281/zenodo.18266001.



\bibliography{ref}

\end{document}



\pagebreak
\listoffigures
\pagebreak

\listoftables
\pagebreak
\newpage

\pagenumbering{arabic}

\pagebreak
\clearpage

\pagebreak
\clearpage





\section{Dataset Descriptions}
\label{sec:data_desc}

Each dataset includes feature columns such as \textit{Release Height}, \textit{Downwind Distance}, \textit{Radionuclide}, and \textit{Stability Category}, with the target variable being the plume shine dose (after $\log_{10}$ transformation during training).

\section{Hyperparameter Configurations}
\label{sed:hyperparam_config}

\begin{table}[h]
    \centering
    \renewcommand\thetable{S\arabic{table}}
    \caption{Optimized hyperparameters used for XGBoost model training. The table summarizes the parameter search space, default settings, and final optimized values obtained through hyperparameter tuning.}
    \label{tab:xgb_params}
    \begin{tabular}{l c c c}
    \hline
    \textbf{Parameter} & \textbf{Distribution / Type} & \textbf{Default} & \textbf{Optimized Value} \\
    \hline
    Objective Function & Fixed & \texttt{reg:squarederror} & \texttt{reg:squarederror} \\
    Evaluation Metric & Fixed & \texttt{rmse} & \texttt{rmse} \\
    Max Tree Depth & Uniform Integer & 6 & 30 \\
    Learning Rate ($\eta$) & Log Uniform & 0.3 & 0.05 \\
    Subsample Ratio & Uniform & 1.0 & 0.5 \\
    Column Subsample by Tree & Uniform & 1.0 & 1.0 \\
    Number of Boosting Rounds & Fixed & 100 & 100 \\
    Early Stopping Rounds & Fixed & 10 & 10 \\
    Device & Categorical & \texttt{cpu} & \texttt{auto (cpu/gpu)} \\
    \hline
    \end{tabular}
\end{table}

\pagebreak
\clearpage

\begin{table}[h]
    \centering
    \renewcommand\thetable{S\arabic{table}}
    \caption{Optimized Random Forest (RF) model hyperparameters used for plume shine dose prediction. The table lists the default parameter values and the corresponding optimized settings adopted during model training.}

    \label{tab:rf_params}
    \begin{tabular}{l c c}
    \hline
    \textbf{Parameter} & \textbf{Default} & \textbf{Optimized Value} \\
    \hline
    Number of Trees ($n_\text{estimators}$) & 100 & 60 \\
    Maximum Tree Depth ($\text{max\_depth}$) & None & 15 \\
    Maximum Features ($\text{max\_features}$) & Auto & 1.0 \\
    Bootstrap Sampling & True & True \\
    Random Seed & --- & 3007 \\
    Number of Jobs & 1 & $-1$ (use all cores) \\
    \hline
    \end{tabular}
\end{table}

\pagebreak
\clearpage

\begin{table}[h]
    \centering
    \renewcommand\thetable{S\arabic{table}}
    \caption{Optimized hyperparameter configuration for the TabNet model obtained using Optuna-based hyperparameter tuning. The table reports the parameter search space, default settings, and final optimized values used for plume shine dose prediction.}

    \label{tab:tabnet_params_appendix}
    \begin{tabular}{l c c c}
    \hline
    \textbf{Parameter} & \textbf{Distribution / Type} & \textbf{Default} & \textbf{Optimized Value} \\
    \hline
    Feature Transformer Dimension ($n_d$) & Discrete Uniform & 8 & 16 \\
    Attention Dimension ($n_a$) & Discrete Uniform & 8 & 16 \\
    Decision Steps ($n_\text{steps}$) & Discrete Uniform & 5 & 10 \\
    Gamma (Sparsity Regularization) & Uniform & 1.0 & 1.0336 \\
    Sparse Regularization ($\lambda_\text{sparse}$) & Log Uniform & $1\times10^{-5}$ & $2.73\times10^{-6}$ \\
    Optimizer & Categorical & \texttt{Adam} & \texttt{Adam} \\
    Learning Rate ($lr$) & Log Uniform & 0.01 & 0.0063 \\
    LR Scheduler & Categorical & None & \texttt{ReduceLROnPlateau} \\
    Scheduler Patience & Integer & 10 & 5 \\
    Scheduler Decay Factor ($\gamma$) & Uniform & 0.95 & 0.9 \\
    Batch Size & Discrete Uniform & 1024 & 2048 \\
    Virtual Batch Size & Discrete Uniform & 128 & 256 \\
    Maximum Epochs & Fixed & 100 & 100 \\
    Early Stopping Patience & Fixed & 10 & 10 \\
    Evaluation Metric & Fixed & RMSE & RMSE \\
    Device & Categorical & \texttt{cpu} & \texttt{auto (cpu/gpu)} \\
    \hline
    \end{tabular}
\end{table}

\pagebreak
\clearpage

\begin{figure}[!htbp]
    \renewcommand\thefigure{S\arabic{figure}}
    \centering
    \includegraphics[width=0.9\linewidth, trim=2pt 40pt 2pt 80pt, clip]{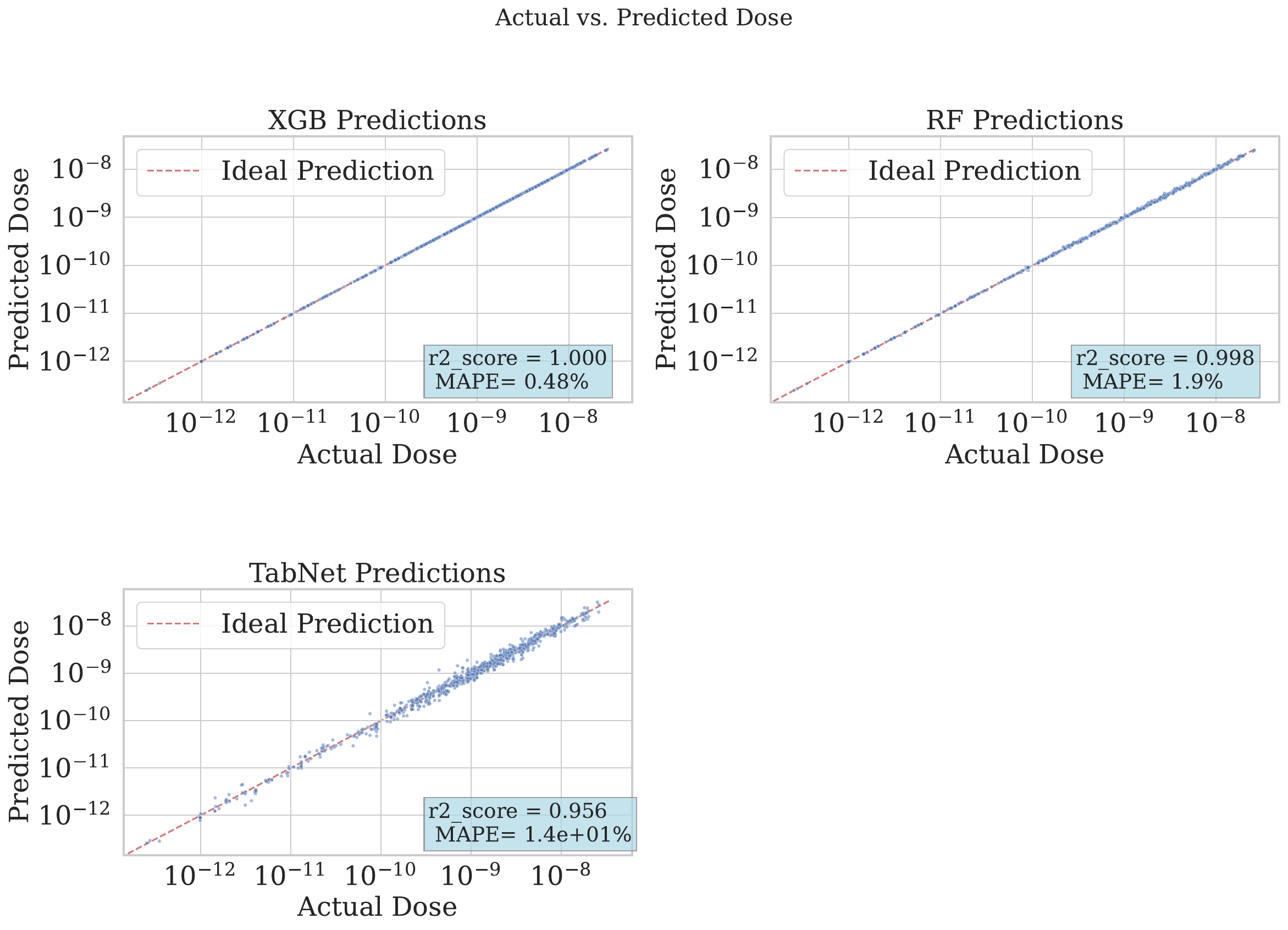}
    \caption{\textbf{Performance of ML/DL models on low-resolution numerical test dataset:} Actual versus predicted plume shine dose obtained using XGBoost, RF and TabNet models trained on $\mathcal{D}_{\mathrm{LR}}^{\mathrm{train}}$ and evaluated on $\mathcal{D}_{\mathrm{LR}}^{\mathrm{test}}$. Results are presented on a logarithmic scale for clarity, with the dashed line indicating the ideal 1:1 prediction. XGBoost exhibits near-perfect agreement with the reference values (R$^{2}$ = 1.000, MAPE = 0.48\%), followed by RF (R$^{2}$ = 0.998, MAPE = 1.9\%), while TabNet shows comparatively larger deviations (R$^{2}$ = 0.956, MAPE $\approx$ 14\%). The comparison highlights the superior predictive capability of tree-based ensemble models for plume shine dose estimation across multiple orders of magnitude.}
    \label{fig:act_pred_lr_lrT}
\end{figure}

\pagebreak
\clearpage

\begin{figure}[!htbp]
    \renewcommand\thefigure{S\arabic{figure}}
    \centering
    \includegraphics[width=0.9\linewidth, trim=2pt 40pt 2pt 80pt, clip]{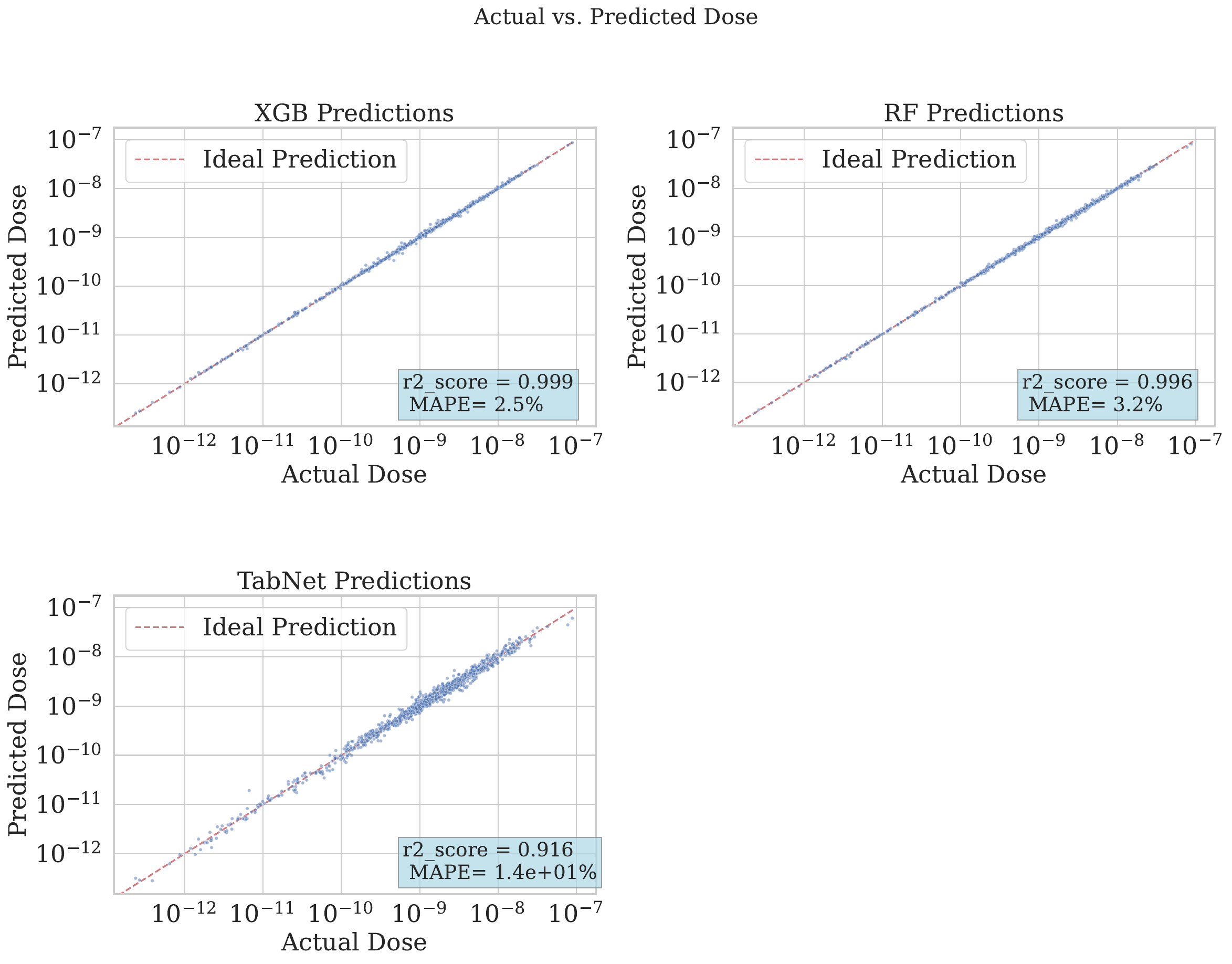}
    \caption{\textbf{Generalization test on interpolated test dataset:} Actual versus predicted plume shine dose obtained using XGBoost, RF and TabNet models trained on $\mathcal{D}{\mathrm{LR}}^{\mathrm{train}}$ and evaluated on $\mathcal{D}{\mathrm{HR}}^{\mathrm{test}}$. Results are presented on a logarithmic scale for clarity, with the dashed line indicating the ideal 1:1 prediction. Compared with evaluation on $\mathcal{D}_{\mathrm{LR}}^{\mathrm{test}}$, all models exhibit a noticeable reduction in predictive accuracy, indicating limited generalization across resolution regimes. XGBoost shows degraded but still relatively strong performance (R$^{2}$ = 0.999, MAPE = 2.5\%), followed by RF (R$^{2}$ = 0.996, MAPE = 3.2\%), while TabNet displays substantially larger deviations (R$^{2}$ = 0.916, MAPE  $\approx$ 14\%). These results suggest that models trained on low-resolution data do not fully generalize to high-resolution distributions, with tree-based ensemble methods remaining comparatively more robust than deep tabular architectures.}
    \label{fig:act_pred_lr_hrT}
\end{figure}

\pagebreak
\clearpage
\begin{figure}[!htbp]
    \renewcommand\thefigure{S\arabic{figure}}
    \centering
    \includegraphics[width=0.9\linewidth, trim=2pt 40pt 2pt 80pt, clip]{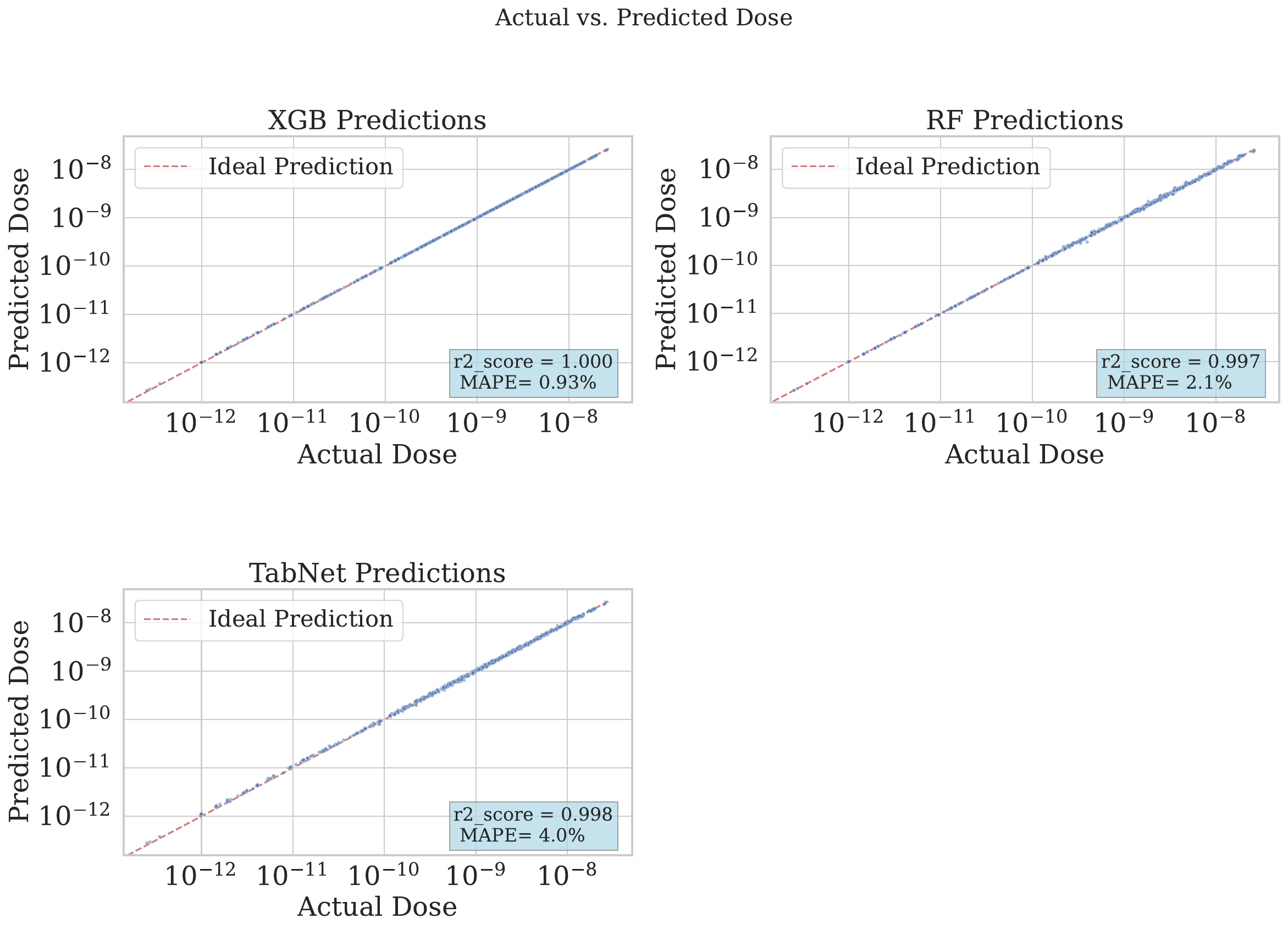}
    \caption{\textbf{Performance of ML/DL models under cross-resolution evaluation:} Actual versus predicted plume shine dose obtained using XGBoost, RF and TabNet models trained on $\mathcal{D}_{\mathrm{HR}}^{\mathrm{train}}$ and evaluated on $\mathcal{D}_{\mathrm{LR}}^{\mathrm{test}}$. Results are presented on a logarithmic scale for clarity, with the dashed line indicating the ideal 1:1 prediction. All models demonstrate strong predictive agreement with the reference values, with XGBoost achieving near-perfect performance (R$^{2}$ = 1.000, MAPE = 0.93\%), followed by RF (R$^{2}$ = 0.997, MAPE = 2.1\%) and TabNet (R$^{2}$ = 0.998, MAPE = 4.0\%). The improved performance relative to the reverse evaluation case indicates that models trained on high-resolution data exhibit better generalization when applied to lower-resolution distributions.}

    \label{fig:act_pred_hr_lrT}
\end{figure}

\pagebreak
\clearpage
\begin{figure}[!htbp]
    \renewcommand\thefigure{S\arabic{figure}}
    \centering
    \includegraphics[width=0.9\linewidth]{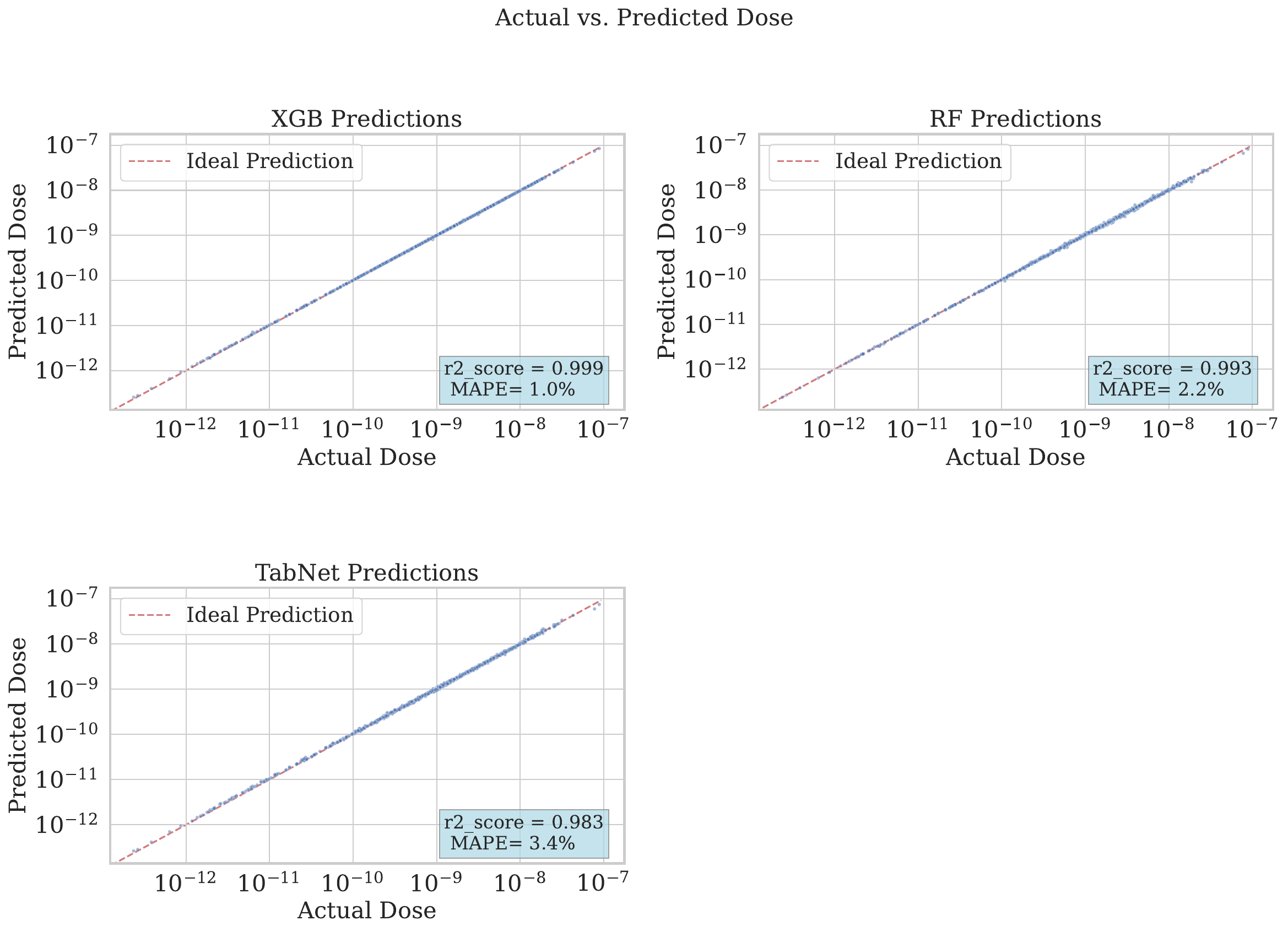}
    \caption{\textbf{Performance of ML/DL models on high-resolution numerical dataset:} Actual versus predicted plume shine dose obtained using XGBoost, RF and TabNet models trained on $\mathcal{D}_{\mathrm{HR}}^{\mathrm{train}}$ and evaluated on $\mathcal{D}_{\mathrm{HR}}^{\mathrm{test}}$. Results are presented on a logarithmic scale for clarity, with the dashed line indicating the ideal 1:1 prediction. All models demonstrate strong predictive agreement with the reference values, with XGBoost achieving the highest accuracy (R$^{2}$ = 0.999, MAPE = 1.0\%), followed by RF (R$^{2}$ = 0.993, MAPE = 2.2\%) and TabNet (R$^{2}$ = 0.983, MAPE = 3.4\%). The results indicate that increasing data resolution improves model learning and reduces prediction uncertainty across multiple orders of magnitude.}
    \label{fig:act_pred_hr_hrT}
\end{figure}